\documentclass[smallextended]{svjour3}       
\smartqed  

\usepackage{lineno,hyperref}
\usepackage{csquotes}
\usepackage{multirow}
\usepackage{array}
\modulolinenumbers[5]

\usepackage{graphicx}
\usepackage{rotating}
\usepackage{enumitem}
\usepackage[caption=false,font=footnotesize]{subfig}
\usepackage{subfig}
\usepackage{todonotes}
\usepackage{verbatim}
\usepackage{xcolor}
\usepackage{tabularx}
\usepackage{booktabs}

\usepackage{array}
\usepackage{ragged2e}
\newcolumntype{P}[1]{>{\RaggedRight\hspace{0pt}}p{#1}}

\newlist{RQL}{enumerate}{1}
\setlist[RQL]{label=RQ-\arabic*:,leftmargin=2.5\parindent}

\newcommand{\ags}{advice-giving system}

\newcommand{\agss}{advice-giving systems}

\newcommand{\RQCHAR}{What are the characteristics of explanations provided to users, in terms of content and presentation?}
\newcommand{\RQAPP}{How are explanations generated?}
\newcommand{\RQEVAL}{How are explanations evaluated?}
\newcommand{\RQCONC}{What are the conclusions of evaluation or foundational studies of explanations?}

\begin{document}

\title{A Systematic Review and Taxonomy of Explanations\\in Decision Support and Recommender Systems}
\titlerunning{A Systematic Review and Taxonomy of Explanations}

\author{Ingrid Nunes \and
       Dietmar Jannach
}


\institute{I. Nunes \at
              Universidade Federal do Rio Grande do Sul (UFRGS), Porto Alegre, Brazil \\
              TU Dortmund, Dortmund, Germany \\
              \email{ingridnunes@inf.ufrgs.br}
           \and
           D. Jannach \at
              TU Dortmund, Dortmund, Germany \\
              \email{dietmar.jannach@tu-dortmund.de}
}

\date{Received: date / Accepted: date}

\maketitle

\begin{abstract}
With the recent advances in the field of artificial intelligence, an increasing number of decision-making tasks are delegated to software systems. A key requirement for the success and adoption of such systems is that users must trust system choices or even fully automated decisions. To achieve this, explanation facilities have been widely investigated as a means of establishing trust in these systems since the early years of expert systems. With today's increasingly sophisticated machine learning algorithms, new challenges in the context of explanations, accountability, and trust towards such systems constantly arise. In this work, we systematically review the literature on explanations in advice-giving systems. This is a family of systems that includes recommender systems, which is one of the most successful classes of advice-giving software in practice. We investigate the purposes of explanations as well as how they are generated, presented to users, and evaluated. As a result, we derive a novel comprehensive taxonomy of aspects to be considered when designing explanation facilities for current and future decision support systems. The taxonomy includes a variety of different facets, such as explanation objective, responsiveness, content and presentation. Moreover, we identified several challenges that remain unaddressed so far, for example related to fine-grained issues associated with the presentation of explanations and how explanation facilities are evaluated.
\keywords{Explanation \and Decision Support System \and Recommender System \and Expert System \and Knowledge-based System \and Systematic Review \and Machine Learning \and Trust \and Artificial Intelligence}
\end{abstract}

\section{Introduction}

In recent years, significant progress has been made in the field of artificial intelligence and, in particular, in the context of machine learning (ML). Learning-based techniques are now embedded in various types of software systems. Features provided in practical applications range from supporting the user while making decisions, e.g.\ in the form of a recommender system, to making decisions fully autonomously, e.g.\ in the form of an automated pricing algorithm. In the future, a constant increase in such intelligent applications is expected, in particular because more types of data become available that can be leveraged by modern ML algorithms. This raises new issues to be taken into account in the development of intelligent systems, such as accountability and ethics~\cite{Banavar:IBM2016:AIEthics}.

A key requirement for the success and practical adoption of such systems in many domains is that users must have confidence in recommendations and automated decisions made by software systems, or at least trust that the given advice is unbiased. This was in fact acknowledged decades ago, when \emph{expert systems}, mainly those to support medical decisions, were popular. Since the early years of expert systems, automatically generated \emph{explanations} have been considered as a fundamental mechanism to increase user trust in suggestions made by the system~\cite{Ye:MISQ1995:ExplanationImpact}. In these systems, provided explanations were often limited to some form of system logging, consisting of a chain of rules that were applied to reach the decision. Nevertheless, such explanations were often hard to understand by non-experts~\cite{187}, thus being used in many cases only to support system debugging.

Today, with modern ML algorithms in place, generating useful or understandable explanations becomes even more challenging, for instance, when the system output is based on a complex artificial neural network~\cite{Samarasinghe:book2006:ANN}. One of the most prominent examples of ML-based applications today are the so-called \emph{recommender systems}~\cite{Jannach:book2010:RecSys}. These systems are employed, e.g., on modern e-commerce sites to help users find relevant items of interest within a larger collection of objects. In the research literature, a number of explanation approaches for recommenders has been proposed~\cite{208,1131,Carenini:AI2006:EvaluativeArguments,Nunes:ECAI2014:Explanations,Bilgic:IUI-BP2005:ExplainingRecommendations}, and existing work has shown that providing explanations can be beneficial for the success of recommenders in different ways, e.g.\ by helping users make better or more informed decisions~\cite{Tintarev:chapter2011:Explanation}.

Therefore, how to explain to the user recommendations or automated decisions made by a software system has been explored in various classes of systems. These include expert systems, knowledge-based systems, decision support systems, and recommender systems, which we collectively refer to as \emph{\agss{}}.\footnote{In general, \emph{\ags{}} is a term that can also include other types of systems, such as conversational agents or autonomous systems. In this paper, however, we use the term to refer only to the four types of systems listed above.} However, despite the considerable amount of research literature in this context, providing adequate explanations remains a challenge. There is, for example, no clear consensus on what constitutes a \emph{good} explanation~\cite{517}. In fact, different types of explanations can impact a user's decision making process in many forms. For instance, explanations can help users make better decisions or persuade them to make one particular choice~\cite{Tintarev:chapter2011:Explanation}. Finally, deriving a \emph{user-tailored} explanation for the output of an algorithm that learns a complex decision function based on various (latent) patterns in the data is challenging without the use of additional domain knowledge~\cite{428}.

Given these challenges, it is important to gather and review the variety of existing efforts that were made in the last decades to be able to design explanation facilities for future intelligent \agss{}. Next-generation explanation facilities are particularly needed when further critical tasks are delegated to software systems, e.g.\ in the domain of robotics or autonomous driving~\cite{678,679}. At the same time, many future \agss{} will need more interactive interfaces for users to give feedback to the system about the appropriateness of the advice made, or to overwrite a decision of the system. In both cases, system-provided explanations can represent a starting point for better \emph{user control}~\cite{Tintarev:chapter2011:Explanation,Jannach:CommACM2016:UserControl,IntRS2017TiiS}.

In this work, we present the results of a \emph{systematic literature review}~\cite{Kitchenham:TechReport2007:SysReviews} on the topic of explanations for \agss{}. We discuss in particular how explanations are generated from an algorithmic perspective as well as what kinds of information are used in this process and presented to the user. Furthermore, we review how researchers evaluated their approaches and what conclusions they reached. Based on these results, we propose a new comprehensive taxonomy of aspects to be considered when designing an explanation facility for \agss{}. The taxonomy includes a variety of different facets, such as explanation objective, responsiveness, content and presentation. Moreover, we identified several challenges that remain unaddressed so far, for example related to fine-grained issues associated with the presentation of explanations and how explanation facilities are evaluated.

Our work is different from previous overview papers on explanations in many ways.\footnote{There is a large amount of surveys that provide an overview of explanations published elsewhere~\cite{Tintarev:chapter2011:Explanation,296,Swartout:chapter1993:ExplanationSurvey,747,Dhaliwal:InfoSysReas1996:ExplanationSurvey,Gregor:MISQ1999:ExplanationSurvey,Moulin:AIReview2002:ExplanationSurvey,Lacave:KER2002:ExplanationSurveyBN,Lacave:KER2004:ExplanationSurvey,Sormo:AIReview2005:ExplanationSurvey,Nakatsu:chapter2006:ExplanationSurvey,1065,609}.} To our knowledge, it is the first \emph{systematic} review on the topic. We initially retrieved 1209 papers in a structured search process and finally included 217 of them in our review based on defined criteria. This systematic approach allowed us to avoid a potential researcher bias, which can be introduced when the selection of the papers that are discussed is not based on a defined and structured process. Moreover, as opposed to some other works, our review does not only focus on one single aspect, such as analysing the different purposes of explanations~\cite{296}. We, in contrast, discuss a variety of aspects, including questions and derived conclusions associated with explanation evaluations. Finally, the comprehensive taxonomy that we put forward at the end of this work is constructed based on the results of a systematic bottom-up approach, i.e.\ its structure is not determined based solely on the authors' expertise in the topic.

\section{Systematic Review Planning}\label{sec:sysRevPlanning}

A systematic review is a type of literature-based research that is characterised by the existence of an exact and transparent specification of a procedure to find, evaluate, and synthesise the results. It includes a careful planning phase, in which goals, research questions, search procedure (including the search string), and inclusion and exclusion criteria are explicitly specified. Key advantages of using such a defined procedure are that it helps to avoid or at least minimise potential researcher biases and supports reproducibility. Such reviews are common in the medical and social sciences and are increasingly adopted in the field of computer science~\cite{Kitchenham:IST2013:SRS}. We next describe the steps that were taken while planning the systematic review. We follow the procedure proposed by Kitchenham and Charters~\cite{Kitchenham:TechReport2007:SysReviews}.

\paragraph{The Need for a Systematic Review}

To our knowledge, no previous work has aimed at providing a comprehensive overview of existing work on explanations in \agss{} using a systematic approach. A number of survey papers exists as mentioned above, but they are either limited to a subjective selection of papers or focused on certain aspects of explanations. Often, existing survey papers simply summarise individual papers and do not consider the developments in the field over time.

\paragraph{Research Goals}

The comprehensive review provided in this paper shall help designers of next-generation \agss{} understand what has already been explored over the last decades in different subfields of computer science. Specifically, we aim to investigate and classify: (i) which forms of explanations were proposed in the literature; (ii) how explanations are generated from an algorithmic perspective; and (iii) how researchers evaluated their approaches and what conclusions they reached. By aggregating the insights obtained from the review in a new multi-faceted taxonomy, we aim to provide an additional aid for designers to understand the various dimensions that one has to potentially consider when designing an explanation facility.

\paragraph{Research Questions}

The specific research questions of the review are consequently as follows.

\begin{RQL}
  \item \label{i:rq1} \RQCHAR
  \item \label{i:rq2} \RQAPP
  \item \label{i:rq3} \RQEVAL
  \item \label{i:rq4} \RQCONC
\end{RQL}

\paragraph{Search Strategies}

To find primary studies that are relevant for our review, we selected the databases presented in Table~\ref{tbl:data-sources}, against which we ran different search queries. Other databases that automatically gather information from difference sources, such as Google Scholar, or allow the addition of non-reviewed papers by their authors, such as arXiv, were left out of the scope of our review. Generally, we assumed that peer-reviewed papers published in computer science are mostly stored in our selected databases. Therefore, these other additional databases would mostly provide duplicated studies. Furthermore, we focused on peer-reviewed work in order to have some evidence regarding the quality of the selected studies.

\begin{table}
  \centering
  \scriptsize
  \caption{Digital Databases used in the Search.}
	\label{tbl:data-sources}
	\begin{tabular}{l l}
		\toprule
		\textbf{Source} & \textbf{URL} \\ \midrule
		ACM Digital Library &\url{http://portal.acm.org} \\
		IEEE Xplore Digital Library & \url{http://ieeexplore.ieee.org} \\
		ScienceDirect  & \url{http://www.sciencedirect.com} \\
		Springer Link & \url{http://link.springer.com} \\
		\bottomrule
  \end{tabular}
\end{table}

\paragraph{Selection Criteria}

Primary studies retrieved from the databases are filtered using a set of criteria. We consider four inclusion criteria (IC) and five exclusion criteria (EC) to select papers associated with the primary studies to be analysed in our review. The inclusion and exclusion criteria are summarised in Table~\ref{tbl:incExcCriteria}. We are interested in four general \emph{types} of studies. Studies involving: (i) the proposal of a \textsc{technique} to generate new forms of explanations (IC-1); (ii) the description of a \textsc{tool} that includes an explanation facility (IC-2); (iii) an \textsc{evaluation} or comparison of one or more forms of explanations  (IC-3); or (iv) a discussion of \textsc{foundational} aspects of explanations (IC-4). Throughout the paper, terms highlighted in small caps are used to refer to these study types.

\begin{table}
  \centering
  \scriptsize
  \caption{Inclusion and Exclusion Criteria.}
	\label{tbl:incExcCriteria}
	\begin{tabularx}{\linewidth}{l X}
		\toprule
		\multicolumn{2}{l}{\textbf{Inclusion Criteria}} \\ \midrule
		\emph{IC-1}: & The paper proposes an explanation generation technique.\\
		\emph{IC-2}: & The paper presents a software application that includes an explanation facility.\\
		\emph{IC-3}: & The paper presents an evaluation of forms of explanations.\\
		\emph{IC-4}: & The paper presents a study that investigates foundations of explanations in \agss{}. \\
		\midrule
		\multicolumn{2}{l}{ \textbf{Exclusion Criteria }} \\
		\midrule
		\emph{EC-1}: & The paper is not written in English.\\
		\emph{EC-2}: & The content of the paper was also published in another, more complete, paper that is already included.\\
		\emph{EC-3}: & The content is not a scientific paper, but an introduction, glossary, etc.\\
		\emph{EC-4}: & We have no access to the full paper.\\
		\emph{EC-5}: & There is a statement in the abstract or content of the paper that explanations are provided, but they are not detailed. \\
		\bottomrule
  \end{tabularx}
\end{table}

The following additional considerations apply regarding the satisfaction of our inclusion criteria. First, in our work, we are only interested in systems in which explanations are provided for \emph{end users}, usually someone that is actually responsible for a final decision. In some studies, visualisations of the outcomes of a data mining process are called explanations as well. Given that such visualisations are designed for data scientists to understand the outputs of the decision algorithms, such works are an example of studies that do not satisfy our inclusion criteria.

Second, we are only interested in studies involving explanations that are related to a specific decision making instance. Generally, an explanation provided by the system can be any form of visual, textual, or multi-modal means to convey additional information to the decision maker \emph{regarding the specific decision to be made}, e.g.\ about the system's reasons to recommend a certain alternative. However, in our review, we do \emph{not} consider approaches in which the system merely provides background knowledge about the domain and this knowledge is independent of the current decision making problem instance. Approaches in which the system displays details about how to interact with user interfaces are also not in the scope of our work.

Third, we consider only scenarios in which there is a limited number of \emph{alternatives}, and the system task is to select one or more of the available choices. In case of multiple selected alternatives, in many cases a ranking is determined by the system. However, a number of papers on decision support systems exist in which the algorithmic task of the system is to compute the single mathematically optimal solution to a given problem. Work on such scenarios, in which there is no set of alternative choices, are also not included in our review. Similarly, studies of explanations regarding the outcomes of mathematical simulations are excluded as well.

With respect to exclusion criteria, we proceeded as follows. In order to obtain the papers in which the studies were published, we first tried to access them through the TU Dortmund network and the Portal de Peri\'{o}dicos CAPES\footnote{\url{http://www.periodicos.capes.gov.br/}}. If the full text was not available, we searched for the paper on the web using: (i) author websites; (ii) Google’s search website; and (iii) repositories of scientific papers, such as Google Scholar and ResearchGate. At the end, only eight studies were excluded because we had no access to the full papers (EC-4).

Finally, there are a number of studies excluded by EC-5 that state that the proposed system has the \emph{potential} to provide explanations, but do not concretely describe how it is done. Examples of studies excluded due to this reason are mainly those in the context of argumentation~\cite{Rahwan:book2009:Argumentation} and studies that focus on transforming a particular output of a decision inference method to a data structure, which is assumed to be easier to explain. For instance, there are approaches that transform artificial neural networks into rules, without detailing how such rules are used to provide explanations to end users.

\section{Systematic Review Execution}\label{sec:sysRevExecution}

The next step in the systematic review process is to execute an appropriate search query against the literature databases. In this section, we provide details of how we constructed the search query and how we ended up with the set of \emph{primary studies} that are considered relevant for our work. The results and insights are then discussed in the remaining sections.

\subsection{Search String Construction}

String construction in systematic reviews is based on a set of terms of interest and their synonyms. Our search string covers two main terms, which both have to appear in the potentially relevant papers. The first term is \emph{explanations}, which is the main topic of our study. In different communities, researchers use alternative terms to refer to explanations, and those were added as synonyms of the term \emph{explanation}. The second term is \emph{decision support system}, which is one of the classes of system that are targets of our review. As synonyms, we consider alternative classes of \agss{} that exist, including in particular knowledge-based and expert systems, as well as recommender systems. For some of them, we included subsets of typically used expressions to refer to such system classes, because they are used with different complementary terms, such as recommendation system, recommendation provider, etc. Therefore, in such cases, we included only the main words as synonyms, because they cover all of the alternatives for the term. The resulting set of synonyms used in our search string is shown in Table~\ref{tbl:synonyms}. The final search string is consequently as follows.

\begin{table}
  \centering
  \scriptsize
  \caption{Terms and their Synonyms.}
	\label{tbl:synonyms}
  \begin{tabularx}{\linewidth}{l X}
    \toprule
    \textbf{Term} & \textbf{Synonyms} \\
		\midrule
    explanation & justification, argumentation \\
    decision support system & decision making, expert system, recommender, recommendation, knowledge-based system, knowledge based system \\
    \bottomrule
  \end{tabularx}
\end{table}

\begin{center}\footnotesize
    \fbox{\parbox[h]{11.5cm}{(explanation OR justification OR argumentation) AND (decision support system OR decision making OR expert system OR recommender OR recommendation OR knowledge-based system OR knowledge based system)}}
\end{center}

The search string was customised to the specific syntax of each of our target databases. In all but one of the cases, we were able to search for the terms in the abstracts of the papers. In the case of the Springer Link database, searching within abstracts was not possible due to API limitations. We thus searched for our terms in the keywords of the papers in this case.

\subsection{Selection of Primary Studies}

After querying the four selected databases on August 12, 2016, we obtained 1209 papers (excluding duplicates) as result. The detailed statistics for each database are given in Table~\ref{tbl:searchResults}.

The relevant primary studies were then selected using a two-step procedure. In the first step, we analysed the title and abstract of each of the 1209 papers. If the title or abstract provided any sort of indication that the paper matched one of the inclusion criteria, the paper was selected to be subsequently analysed in detail. We also pre-classified each paper according to one of the inclusion criteria, i.e.\ we identified a preliminary study type.

In the second step, the full text of each paper was retrieved, if available, and analysed. Then, we checked each of the exclusion criteria and re-evaluated whether the paper truly satisfied one of the inclusion criteria. As result, some papers were discarded or associated with a different inclusion criterion.

Each abstract and paper was analysed by a single researcher, strictly following the specified protocol. When it was not fully clear whether a certain criterion was satisfied, i.e.\ border cases, the opinion of a second researcher was requested in order to minimise the potential researcher bias.

\begin{table}
  \centering
  \scriptsize
  \caption{Search Results by Source.}\label{tbl:searchResults}
  \begin{tabular}{l r}
		\toprule
    \textbf{Source} & \textbf{Number of Studies} \\
		\midrule
    ACM Digital Library & 278 \\
    IEEE Xplore Digital Library & 469 \\
    ScienceDirect & 385 \\
    Springer Link & 92 \\
		\midrule
    Duplicates & 15 \\
		\midrule
    \textbf{Total (including duplicates)} & 1224 \\
    \textbf{Total (excluding duplicates)} & 1209 \\
    \bottomrule
  \end{tabular}
\end{table}

Following this procedure, we ended up with a total of \emph{217 primary studies}. The detailed statistics about the paper selection process are presented in Table~\ref{tbl:filteringResults}.  The column headings correspond to the applicable inclusion criteria, and rows with labels starting with \emph{EC} show how many papers satisfied each exclusion criterion. The full list of papers that was finally considered is shown in Table~\ref{tbl:selectedStudies}.\footnote{Further details about which inclusion and exclusion criterion was fulfilled by each individual paper, along with the detailed results of our analysis, can be found online at: \url{http://inf.ufrgs.br/prosoft/resources/2017/umuai-sr-explanations}.}


\begin{table}
  \centering
  \scriptsize
  \caption{Selection of Primary Studies according to Inclusion and Exclusion Criteria.}
	\label{tbl:filteringResults}
  \begin{tabular}{l r r r r r}
    \toprule
    \textbf{Step / Criterion}      & \textbf{Technique}& \textbf{Tool} & \textbf{Evaluation} & \textbf{Foundational} & \textbf{Total} \\
		\midrule
		Abstract Analysis	         	  & 192								& 161 					& 30									 & 3 									& 386\\
		Full Analysis	                & 135								& 139						& 23									 & 5 									& 302\\
		\midrule
		EC-01							           	&  									& 1 						&  										 &   									& 1  \\
		EC-02								          & 10    						& 2 						& 1										 &   									& 13 \\
		EC-03								          & 2									& 5 						 &  										&   									& 7  \\
		EC-04							          	& 3    							& 5 						&  										 &   									& 8  \\
		EC-05								          & 20								& 37						&  										 &   									& 57 \\
		\midrule
		\textbf{Selected Studies}    	& \textbf{101}     	& \textbf{89}	  & \textbf{22}					& \textbf{5} 					 & \textbf{217}\\
		\bottomrule
		\multicolumn{6}{l}{One \textsc{technique} paper satisfies both EC-02 and EC-04.} \\
  \end{tabular}
\end{table}

\begin{table}[t]
  \centering
  \scriptsize
  \caption{Selected Primary Studies.}
	\label{tbl:selectedStudies}
  \begin{tabularx}{\linewidth}{X}
    \toprule
    \textbf{Approach} \\
		\midrule
		 \cite{5,31,65,71,77,81,84,86,110,111,115,137,141,151,187,189,248,334,339,356,377,381,386,389,415,422,428,432,437,438,454,455,456,491,508,509,511,528,532,533,536,553,566,576,581,596,605,606,614,634,648,678,705,710,717,735,744,748,751,759,784,806,813,831,834,838,851,852,853,866,878,894,897,904,905,920,930,943,951,952,959,975,976,1002,1017,1023,1028,1037,1045,1048,1088,1093,1103,1128,1133,1141,1164,1173,1185,1188,1189} \\
		\midrule
		\textbf{Tool} \\
		\midrule
		 \cite{11,69,72,75,83,85,89,102,103,104,122,123,135,136,168,169,177,181,183,195,199,212,256,257,260,262,263,310,315,338,350,400,423,460,547,579,602,630,711,718,721,727,729,732,741,745,753,757,761,762,764,767,776,779,782,783,785,791,810,811,819,832,836,841,845,857,868,871,874,886,898,910,911,916,919,928,934,950,973,1019,1034,1047,1083,1111,1124,1126,1135,1170,1197} \\ \midrule
		\textbf{Evaluation} \\
		\midrule
		\cite{208,272,487,530,550,598,633,651,655,679,789,817,860,861,872,955,998,1058,1073,1076,1131,1183} \\
		\midrule
		\textbf{Foundation} \\
		\midrule
		\cite{303,517,823,936,1167} \\
		\bottomrule
  \end{tabularx}
\end{table}

\section{Results}\label{sec:sysRevResults}

In this section, we summarise the insights that we obtained by analysing the 217 selected primary studies with respect to our research questions.\footnote{The interested reader can find the detailed analysis of each of the studies on the following link: \url{http://inf.ufrgs.br/prosoft/resources/sr-explanations}.} These insights are used later on to develop a comprehensive taxonomy of aspects to consider when designing future explanation facilities.

\subsection{Historical Developments}

\begin{figure}
	 \centering
     \subfloat[Number of Primary Studies.\label{fig:type-year-amount}]{%
       \includegraphics[width=0.49\linewidth]{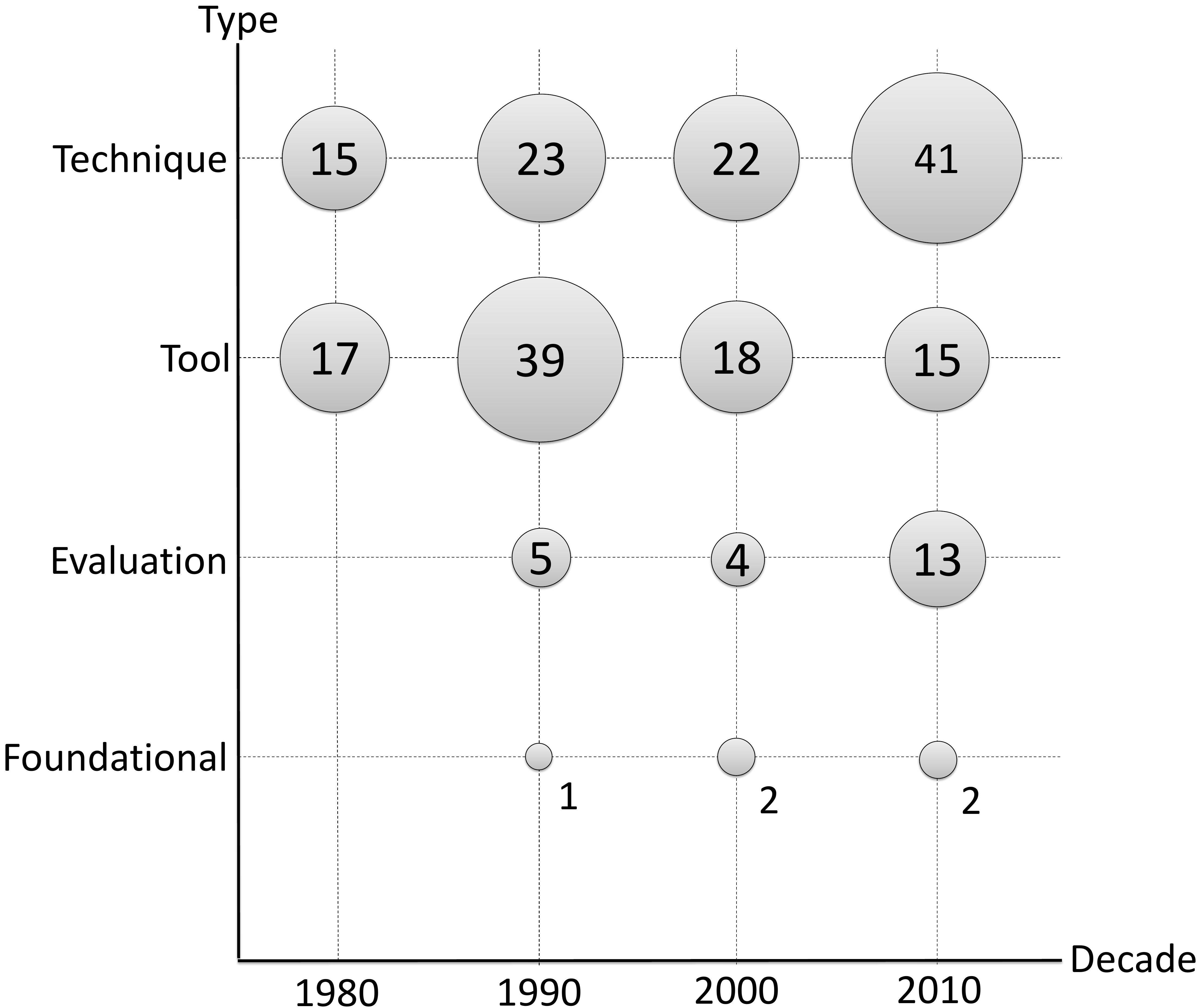}
     }
     \subfloat[Percentage of Primary Studies.\label{fig:type-year-percentage}]{%
       \includegraphics[width=0.49\linewidth]{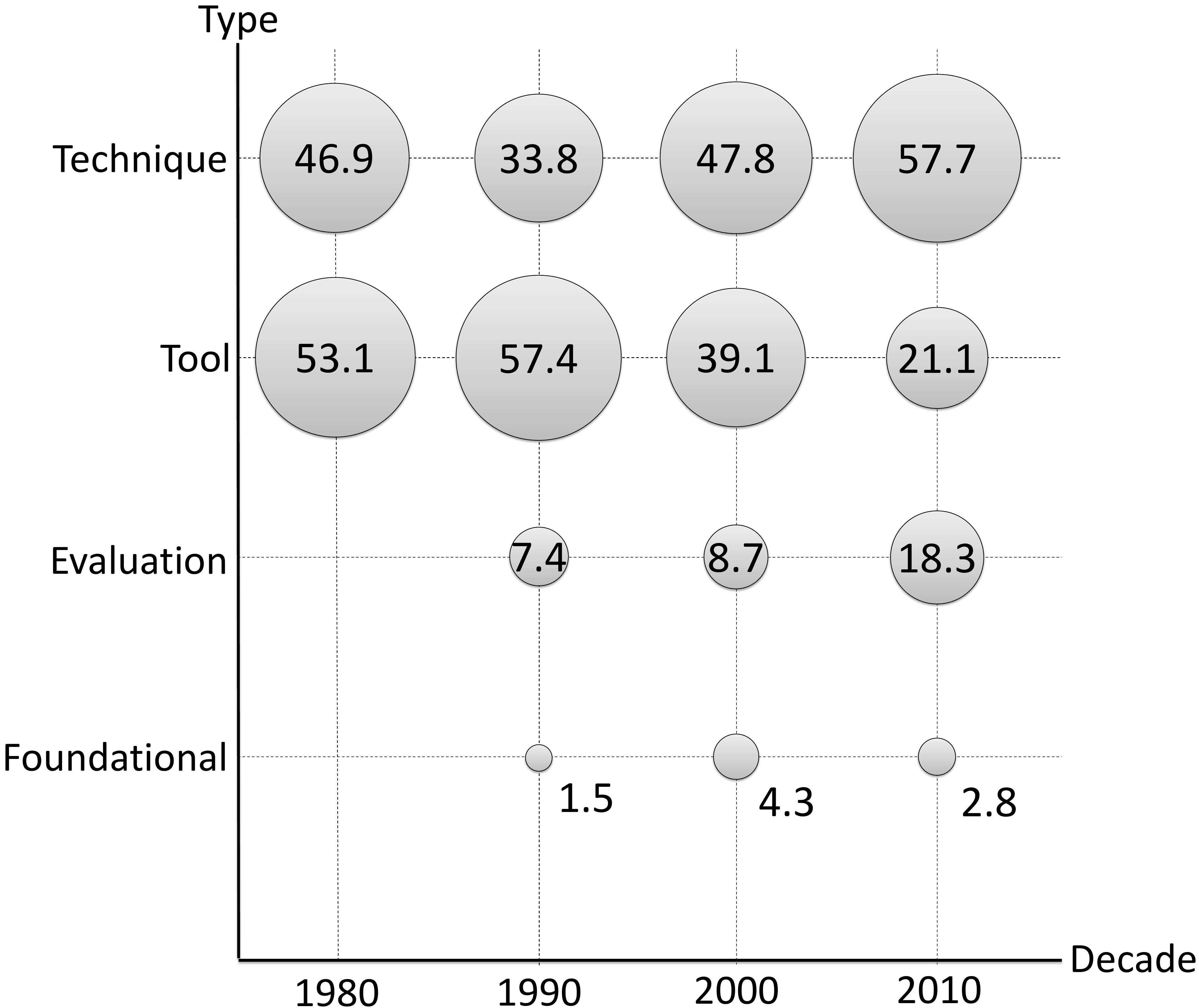}
     }
     \caption{Primary Studies: Number of Studies by Type per Decade.}
     \label{fig:type-year}
\end{figure}

Before focusing on our research questions, we provide an overview of the historical developments in the research field of explanations by analysing the papers associated with the examined studies in terms of type and publication date. Figure~\ref{fig:type-year-amount} shows how many papers of each type were published \emph{per decade} in absolute numbers. Figure~\ref{fig:type-year-percentage} is based on the same data, but it reports percentages within the corresponding decade instead of absolute numbers. The following main observations can be made.

\begin{itemize}

	\item Generally, the total number of published papers in the field increases. We emphasise that the last decade (2010--present) corresponds to only about 6.5 years.\footnote{Note that more papers were published in computer science in general over time.}
	
	\item Papers on \textsc{tools} with explanation facilities were much more common in the past, perhaps because such papers were more often considered as research contributions at that time. In exchange, an increase over time can be observed in the number of papers related to the proposal of new \textsc{techniques}.
	
	\item The empirical \textsc{evaluation} of explanations received much more attention in the recent past. This is an indication that the field achieved a higher level of maturity due to the progress made in terms of research methodology.
	
	\item Papers on \textsc{foundational} aspects of explanations are very scarce.

\end{itemize}

From a historical perspective, it seems that research on the topic of explanations reached some plateau in the 1990s. In the 2000s, we see a stagnation, but a considerable increase again in the past few years. We attribute the observed stagnation in the 1990s to the declining role of knowledge-based systems at that time. In some areas of decision support systems, and particularly in the field of recommender systems, ML-based approaches became predominant in the 2000s and researchers focused more on determining the \emph{right} recommendations than on the provision of explanations.

A measurable increase can be seen in the number of published research papers on the topic in the past few years, indicating the importance and timeliness of the topic. Furthermore, we expect more works in the future caused by two recent trends in computer science. First, explanations of the behaviour of a software system that are directed to the end user are clearly a key ingredient for modern \emph{human-centric computing} approaches~\cite{Jaimes:Computer2007:HCC}. Second, as discussed, an increasing number of tasks are expected to be delegated to automated software systems that are based on modern ML technology in the future. However, to be accepted by end users, the suggestions made by \agss{} must be perceived to be fair and transparent in many application domain, and explanations are key to this.\footnote{See \url{http://www.fatml.org} for a recent workshop series on fair, accountable, and transparent machine learning approaches.}

\subsection{\ref{i:rq1} \RQCHAR}  \label{sec:rq1}

To answer our first research question, we used the analysis method described next. Primary studies analysed in the context of this research question are those that proposed forms of explanations, consisting of \textsc{techniques} and \textsc{tools}. Together, such types of studies are referred to as (explanation generation) approaches. We discuss the obtained results in terms of the content and presentation of explanations.

\subsubsection{Analysis Method}
\label{sec:research-method-grounded-theory}

We followed principles from \emph{grounded theory}~\cite{Glaser:book1992:GroundedTheory} to investigate this question in a systematic and unbiased way. Following these principles, we iterated through all primary studies and labelled their proposed explanations with \emph{codes} (in grounded theory terminology) that captured key ideas associated with the explanation content. For example, consider an explanation approach that has the following system output: \enquote{The recommended alternative has A and B as positives aspects, even though it has C as a negative aspect.} Explanations of this type, in which the features of different alternatives are contrasted, were labelled with the code \emph{Pros and Cons}.

The inspection of all studies led us to a first set of codes. This preliminary set was then analysed, in order to merge codes that represent the same underlying idea. For instance, there is an explanation that organises the suggested alternatives in groups to highlight the trade-off relationships between certain features~\cite{1017}. This and similar approaches were initially labelled with the code \emph{Trade-off}. As both \emph{Pros and Cons} and \emph{Trade-off} represent the same underlying idea of explaining the alternative options, these codes were merged.

At the end of the process, each form of explanation proposed in the considered studies was labelled with one or more codes related to \emph{what} kind of information is presented. After merging the codes, we ended up with 26 labels. A similar method was adopted regarding \emph{how} the information is presented.

\subsubsection{Explanation Content}

\begin{figure}
	 \centering
      \includegraphics[width=\linewidth]{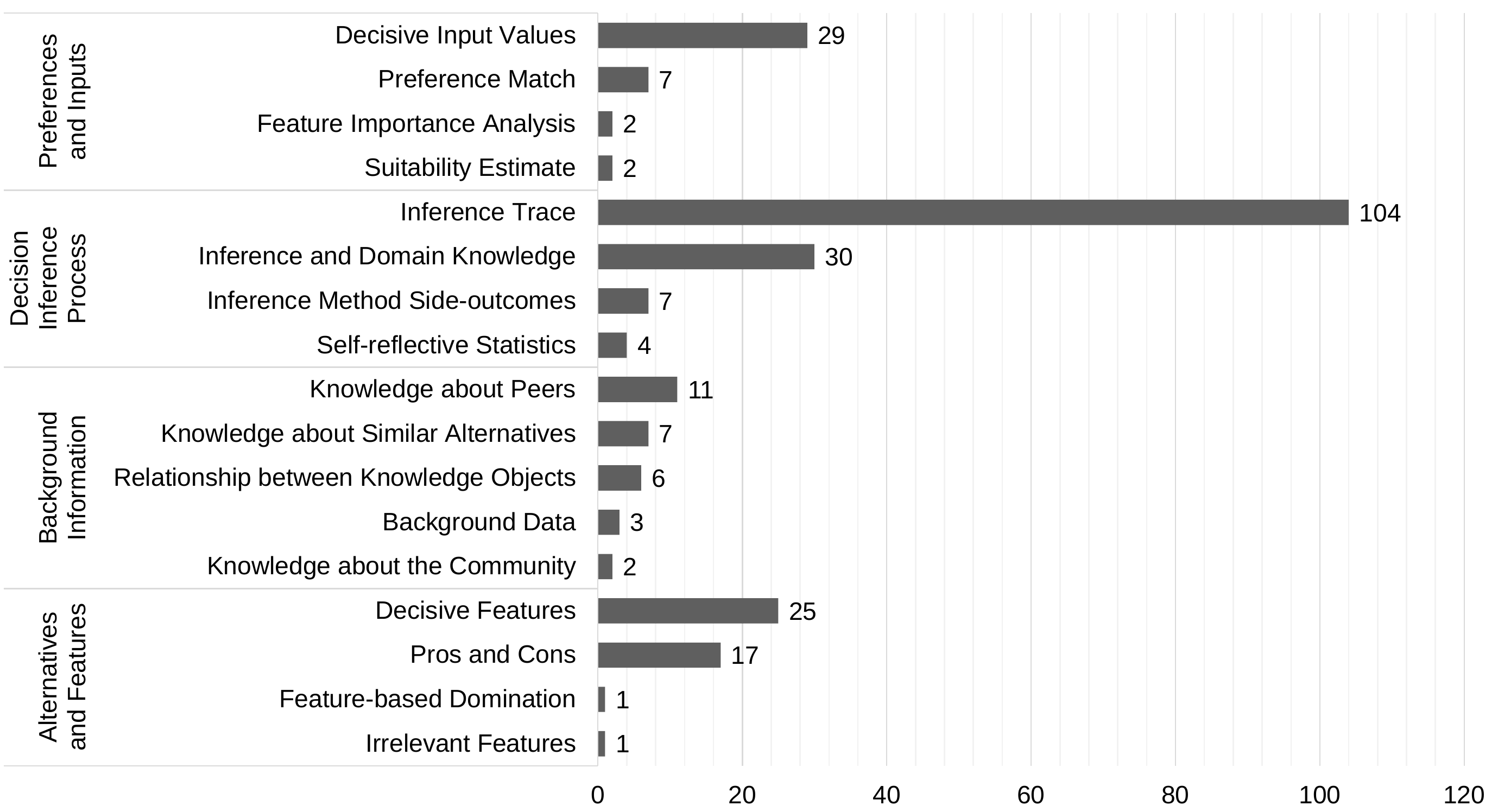}
     \caption{Occurrence of Codes related to Explanation Content.}
     \label{fig:labels-content}
\end{figure}

From the 26 codes, 17 refer to the type of information that was displayed, while the 9 remaining codes are about general observations regarding content, such as cases where explanations are context-tailored. The content-related codes are described in detail in Table~\ref{tbl:orientationLabels}, while their occurrence frequencies are shown in Figure~\ref{fig:labels-content}. The different types of information presented in the explanations can be organised in four main groups, as indicated in Table~\ref{tbl:orientationLabels} and detailed as follows.

\begin{table}
	\newcommand{\tabcat}[4]{\multirow{#1}{*}[#2]{\renewcommand\baselinestretch{1}\selectfont \begin{sideways}\parbox[c]{#3}{\centering #4}\end{sideways}}}
	\caption{Codes related to the Type of Information Conveyed in Explanations.}
	\scriptsize
	\renewcommand{\tabcolsep}{0.5mm}
	\centering
	\begin{tabularx}{\linewidth}{ c @{\hskip 5pt} >{\setlength{\baselineskip}{0.7\baselineskip}} P{2.8cm} @{\hskip 5pt} >{\setlength{\baselineskip}{0.7\baselineskip}}X  }	
		\toprule
		&\textbf{Label} & \textbf{Description} \\
		\midrule
		\tabcat{4}{0.95cm}{4cm}{User Preferences\\and Input}
			& Decisive Input Values & Indication of the inputs that determined the resulting advice. \\ \cline{2-3}
			& Preference Match & Provision of information about which of the user preferences and constraints are fulfilled by the suggested alternative. \\ \cline{2-3}
			& Feature Importance Analysis & Justification of the advice in terms of the relative importance of features, e.g.\ by showing that changing feature weights would cause the selected alternative to be different. \\ \cline{2-3}
			& Suitability Estimate & Indication of how the system believes that the user would evaluate the suggested alternative, e.g.\ by showing a predicted rating. \\
		\midrule
		\midrule
		\tabcat{5}{0.6cm}{4cm}{Decision Inference\\Process}
			& Inference Trace  & Provision of details of the reasoning steps that led to the suggested alternative, e.g.\ a chain of triggered inference rules. \\ \cline{2-3}
			& Inference and Domain Knowledge & Provision of information about the decision domain or process, e.g.\ about the main logic of the inference algorithm. For example: \enquote{We suggest this option because similar users liked it.} \\ \cline{2-3}
			& Decision Method{\par}Side-outcomes & Provision of algorithm-specific outcomes of the internal inference process, e.g.\ a calculated number that expresses the system's confidence. \\ \cline{2-3}
			& Self-reflective{\par}Statistics & Provision of facts regarding the system's performance, e.g.\ by informing the user how many times the system made decision suggestions in the past that were accepted. \\
		\midrule
		\midrule
		\tabcat{4}{0.1cm}{4cm}{Background and\\Complementary Information}
			& Knowledge about Peers  & Provision of information about the preferences of related users, e.g.\ ratings given to a suggested alternative by social friends. \\ \cline{2-3}
			& Knowledge about Similar Alternatives & Indication of similar alternatives that were an appropriate (system's or user's) decision in a similar context in the past, e.g.\ items that the user or related peers showed interest in. \\ \cline{2-3}
			& Relationship between Knowledge Objects & Provision of information about the relationship between features, or features and users. This can be done, for example, in the form of a directed acyclic graph representing a causal network. \\ \cline{2-3}
			& Background Data & Provision of (external) background data specific to the current problem instance, e.g.\ data derived from processing posts in a social network, which were considered in the decision inference process. \\ \cline{2-3}
			& Knowledge about the Community & Provision of information that supports the decision based on the behaviour and preferences of a community, e.g.\ showing the general popularity of the proposed alternative. \\
		\midrule
		\midrule
		\tabcat{5}{0.6cm}{4cm}{Alternatives\\and their Features}	
			& Decisive Features & Indication of the features of the alternative that are key to the decision. \\ \cline{2-3}
			& Pros and Cons & Indication of the key positive and negative features of the alternative. \\ \cline{2-3}
			& Feature-based Domination & Justification of a decision in terms of the dominance relationship between two alternatives, e.g.\ by showing that an alternative is not selected because it is dominated by another. \\ \cline{2-3}
			& Irrelevant Features & Indication of features that are irrelevant for the decision, typically when the values of such features in the suggested alternative are not considered good. \\
		\bottomrule
	\end{tabularx}
	\label{tbl:orientationLabels}
\end{table}%

\begin{description}

\item [User Preferences and Inputs.] A possible way to explain a system's suggestion is to provide users with explanatory information that is related to the provided inputs. The explanation can, for example, indicate (i) which of the user preferences and constraints were fulfilled and which were not, (ii) to what extent the system believes that the recommended alternative is appropriate given the stated preferences, or (iii) which inputs were the most decisive when determining the suggestion.

\item [Decision Inference Process.] Providing information about the inference process of a specific decision problem (e.g.\ in the form of traces) was the most common approach in classical expert systems. Some explanations only provide the general logic of the system's internal inference process; others mention system confidence in the suggestion or the success rate in past decision making situations.

\item [Background and Complementary Information.] A reduced amount of explanations provide additional background information that is specific to the given decision making instance. Various types of background and complementary information were identified. Explanations can, for example, provide more information about the knowledge sources that were used in the inference process or how relevant entities in the knowledge base are interrelated. They can also refer to past suggestions or user choices in similar situations, or mention which users liked the suggested alternative.

\item [Alternatives and their Features.] A common approach in the literature is to explain the system's suggestion by analysing the features of the alternatives. Some explanations consist of lists of features, pro and con, for each alternative; others refer to dominance relationships based on the features, but most explanations show which features were decisive in the inference process.

\end{description}

We made the following additional observations regarding orthogonal aspects when analysing which kind of information is conveyed to the user within the explanations.

\paragraph{Baselines and Multiple Alternatives} First, in most cases the provided explanations focus solely on one single (recommended) alternative. For example, such explanations contain details about the features of the alternative, describe in which ways it is suitable for the user, or how the decision was made. However, there are some approaches that use other alternatives as a \emph{baseline} for comparison. We observed two forms of including baselines in such a comparison. One option is to use one single alternative (e.g.\ the second best choice), as done in~\cite{951,1048}. As an example, the provided explanation can detail in which ways the best alternative is favourable over the second best option. A different approach is to contrast one alternative with a \emph{set} of other options~\cite{256,432,764}. In this case, the explanation highlights why the best alternative is generally better than the group of other alternatives, e.g.\ in terms of specific features. Finally, there are two cases~\cite{834,1017} in which the explanation does not refer to one single best alternative as a reference point to compare other options with, but to a \emph{group} of alternatives (equally suitable regarding a considered aspect) or groups of rules that were used in the inference process.

In one of these studies~\cite{1017}, a key goal is to educate users about the trade-off among options by means of explanation interfaces. This can be achieved by these explicitly provided explanations and also by providing interactive decision making support, such as by means of dynamic critiquing~\cite{McCarthy:IUI2005:DynamicCritiquing}.

\paragraph{Context-tailored Explanations} Which information an explanation should provide to the user can depend on various factors, including the expertise or interests of users, or their current situational context. We identified 16 primary studies in which explanations are tailored to the current situation in different forms, e.g.\ by using different levels of detail. Moreover, \emph{group decisions} can be seen as a very specific context. In our review, we found only one single approach~\cite{911} that focuses on explaining decision suggestions that were made for a group of users.

\paragraph{External Sources of Explanation Content} As discussed, some explanations provide information associated with background knowledge, but such knowledge is almost always associated with the decision inference process. Four approaches~\cite{596,606,634,648} in the e-commerce domain exploited external sources of information, namely \emph{product reviews}.

\paragraph{Interactive Explanations} In some approaches, the explanations provided by the system represent a starting point for further user interactions, e.g.\ asking the user for additional input. The common types of questions associated with user interaction are: (i) \emph{what-if} (what the output would be if alternative input data were provided); (ii) \emph{why} (why the system is asking for a particular input); and (iii) \emph{why-not} (why the system has not provided a given output). These questions were addressed in 10, 37, and 13 studies, respectively. These three types of questions, together with the \emph{how}-question (how the system reached a certain conclusion), are those typically addressed in expert systems. Explanations that address the \emph{how}-question appear in Table~\ref{tbl:orientationLabels}, mainly as inference traces. In addition, some explanations given as answers to this question trace the path from a decision to the given user input. As result, they only report the input that actually led to the decision.

\subsubsection{Explanation Presentation}

\begin{figure}
	 \centering
   \includegraphics[width=\linewidth]{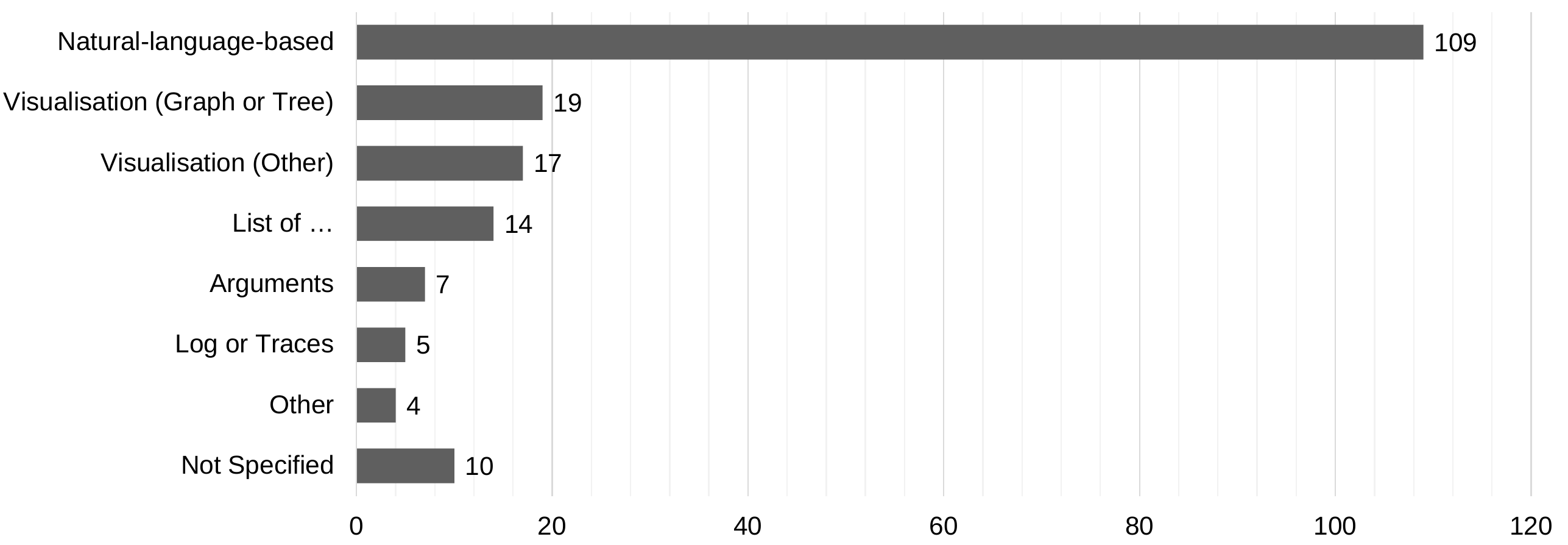}
   \caption{Occurrence of Codes related to Explanation Presentation.}
   \label{fig:labels-presentation}
\end{figure}

Now that we have discussed the content of explanations, we proceed to how they are \emph{presented} to users. Codes were also used in this analysis. As result of the categorisation of the different ways of how explanations are presented, we identified 8 presentation codes, which are shown along with their occurrence frequencies in Figure~\ref{fig:labels-presentation}.

The most frequent code is by far \emph{natural-language-based}, i.e.\ most of the approaches use a text format to display explanations to the user. Note that we also used this code to label explanations that are based on pre-defined templates, which were, for example, instantiated with lists of features before they were presented to the user. Alternatively, displaying simple lists of various things (e.g.\ features, users, alternatives, past cases, conclusions derived in the decision process) was chosen as a presentation form by a number of studies. Different forms of visualisations, e.g.\ in the form of graphs, are also quite common means to convey the explanatory information to the user. Only a smaller number of studies presented inference traces (or other forms of logs) to the user as a final explanation presentation format. Finally, some studies structure the explanation as an argument, typically in the form of the Toulmin's argument structure~\cite{Toulmin:book2003:Arguments}.

There are four studies that used unusual forms of presentations, and we assigned them to the group called \emph{Other}. These are the types of outputs in this group: (i) audio~\cite{83}, which uses voice as output; (ii) highlighting~\cite{423}, which presents an alternative with highlighted aspects; (iii) query results~\cite{813}, when the explanation is the result of a database query; and (iv) OWL (Ontology Web Language)~\cite{1023}, when explanations are given using this technical language (to be further processed for presentation to the end user). Studies that do not specify how the explanations are presented to the user were assigned to the group \emph{Not Specified}.

\subsection{\ref{i:rq2} \RQAPP} \label{sec:rq2}

Having discussed what is actually presented as explanations to end users, we now proceed to the investigation of \emph{how} they are generated in the proposed approaches. Specifically, we are interested in the inner workings of processes that take the outcomes of a decision inference method (that was used to determine the suggested alternative) to produce what is finally shown as an explanation.

\subsubsection{Explanation Generation Process}

We observed that most of the studies investigated in our review do not provide many details about this process. The reason is that in most cases the explanation generation process is closely tied to the underlying decision inference method and the data that is used to determine the suggested alternative. If, for example, the underlying inference method is rule-based, the explanation presented to the user might consist of a set natural language representations of the rules that were triggered. In many of these cases, no further information is provided regarding whether any additional processing was needed to generate what is finally presented to the user in one of the different forms shown in the previous section.

Only a few studies implemented more complicated explanation generation processes. In particular, when the underlying inference method is based on multi-criteria decision theory (MAUT)~\cite{Keeney:book1976:MAUT},  to derive an explanation for the user, specific algorithms are often provided to analyse user preferences with respect to the features of the alternatives (e.g.~\cite{648,831,1048}). Approaches based on artificial neural networks (e.g.~\cite{783,836,836,853}) also often specify algorithms to extract rules from the network to generate the explanations.\footnote{We remind the reader that approaches that simply provide a rule extraction algorithm without detailing how the explanations are provided to the end user are excluded from our review as specified in our inclusion and exclusion criteria.}

Although there is limited information regarding the explanation generation process in most of the approaches, we identified three factors that strongly influence the selection of what sort of explanation will be provided. In a few studies, the \emph{application domain} plays a key role in this process. 18 (out of 101, or 17.8\%) explanation generation \textsc{techniques} are domain-specific, that is, their proposals are focused on, and possibly tailored to, a particular domain. Domain-specific approaches can be found in the recent past mostly in the fields of \emph{Computing \& Robotics} and \emph{Media Recommendation}. They exploit information types that are only available in a certain domain to produce richer explanations. The approach proposed by Briguez et al.~\cite{1128} is an example of such a work, in which the authors identified specific argument structures that can be used in the media (movie) recommendation domain. However, from \textsc{techniques} that focus on a particular domain, many do not exploit any domain specificities. Domain specificity is not analysed in studies involving \textsc{tools}, because they are always developed with a focus on a particular domain.

In addition to the application domain, we identified two other key drivers of the explanation generation process. The first is the \emph{purpose} for which explanations are provided in an \ags{} and, second, the adopted \emph{underlying decision inference method}, as briefly discussed above. These are further discussed in the next sections.

\subsubsection{Purpose of Explanations}
\label{sec:purposes}

Previous surveys highlighted the importance of considering the intended purpose of explanations when designing an explanation facility~\cite{296,Buchnan:chapter1984:ExplanationSurvey}. The intended purpose, which can be, for example, to persuade users to accept a suggested alternative, should determine what information should be conveyed to the user. However, most of the studies analysed in our review do not explicitly state their purpose or, more specifically, their \emph{intended} purpose~\cite{Friedrich:AIMag2011:ExplanationTaxonomy}.

Therefore, we adopted an analysis method, in which we searched in \textsc{technique} and \textsc{tool} studies for sentences that indicate the underlying explanation purpose of the study. To categorise the different studies, we used the list of possible purposes proposed by Tintarev and Masthoff~\cite{296} as a basis. We then extended this list with additional categories that we found during the analysis and specifically included categories that were identified in an earlier work on explanations by Buchnan and Shortliffe~\cite{Buchnan:chapter1984:ExplanationSurvey}.

The resulting list of explanation purposes found in the studies is shown in Table~\ref{tbl:explanationPurpose}. Some of the mentioned explanation purposes from the two reference works are in fact similar, but the authors used different labels, e.g.\ transparency \emph{vs.\ }understanding or trust \emph{vs.\ }acceptance. In these cases, we used the label that is used in the more recent reference work.

\begin{table}
	\scriptsize
	\centering
	\caption{Explanation Purposes Identified in Primary Studies (based on~\cite{296,Buchnan:chapter1984:ExplanationSurvey}).}
	\label{tbl:explanationPurpose}
	\begin{tabular}{l l l}
			\toprule
			\textbf{Purpose} & \textbf{Source} & \textbf{Description} \\
			\midrule
			Transparency & \cite{296,Buchnan:chapter1984:ExplanationSurvey} & Explain how the system works \\
			Effectiveness & \cite{296} & Help users make good decisions \\
			Trust & \cite{296,Buchnan:chapter1984:ExplanationSurvey} & Increase users' confidence in the system \\
			Persuasiveness & \cite{296,Buchnan:chapter1984:ExplanationSurvey} & Convince users to try or buy \\
			Satisfaction & \cite{296} & Increase the ease of use or enjoyment \\
			Education & \cite{Buchnan:chapter1984:ExplanationSurvey} & Allow users to learn something from the system \\
			Scrutability & \cite{296} & Allow users to tell the system it is wrong \\
			Efficiency & \cite{296} & Help users make decisions faster \\
			Debugging & \cite{Buchnan:chapter1984:ExplanationSurvey} & Allows users to identify that there are defects in the system \\
			\bottomrule
	\end{tabular}
\end{table}%

\begin{figure}
	\centering
	\includegraphics[width=\linewidth]{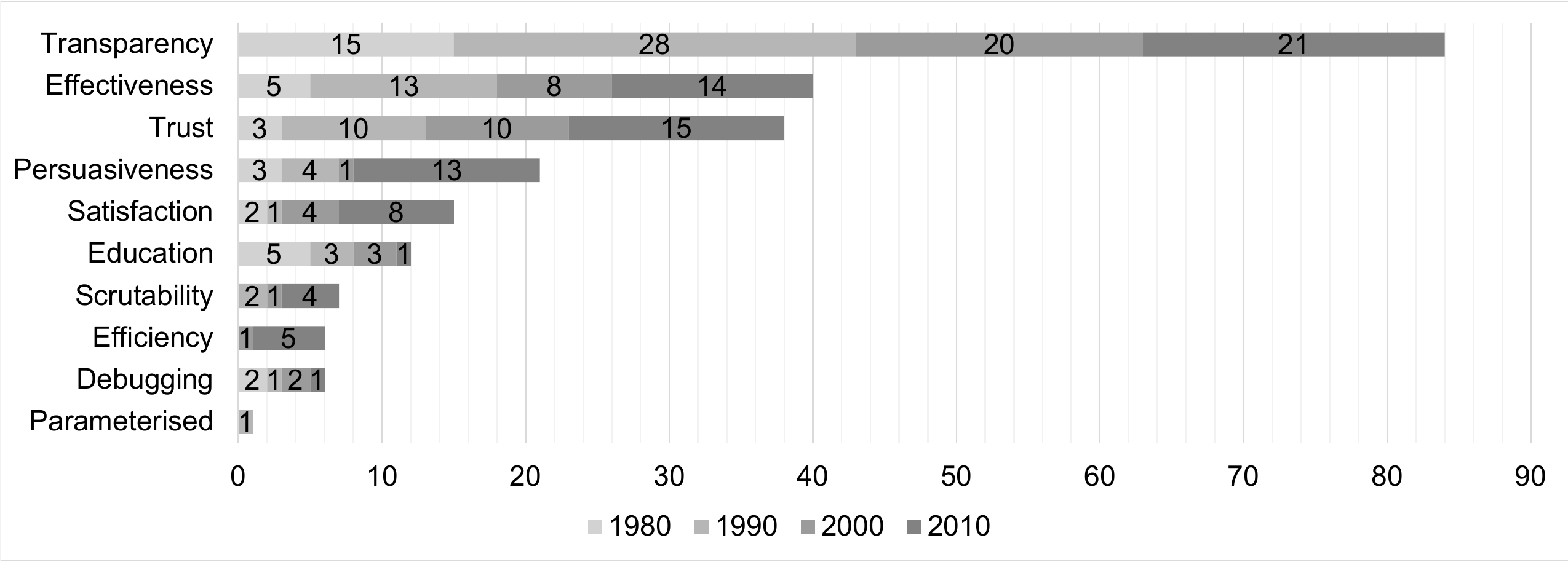}
	\caption{Purpose Analysis: Number of \textsc{Technique} or \textsc{Tool} Studies per Purpose.}
	\label{fig:purpose}
\end{figure}

Figure~\ref{fig:purpose} shows how often each intended purpose is in the focus of the investigation in our primary studies. Each study can be associated with more than one purpose. The majority of the studies (124, or 65\%) are, however, concerned only with one single purpose. From the studies associated with multiple purposes, only one study~\cite{735} explicitly states five of them (maximum). In one case~\cite{845}, the purpose is a parameter of the explanation generation process (and classified as \emph{parameterised} in Figure~\ref{fig:purpose}). In 11\% of the cases (21 studies), we could not identify a single sentence within the respective paper that indicates the purpose of the explanations.

As can be seen in Figure~\ref{fig:purpose}, in which we also indicate the decade in which the respective studies were published, the most common explanation purpose is to provide \emph{transparency}, i.e.\ to explain how the system ended up with the suggested alternative(s). The provided explanations in these studies focused on exposing the system's inference process in order to make the recommended decision understandable. The authors of one study~\cite{5} argued that this aspect is particularly critical because ``\emph{[the] human user bears the ultimate responsibility for action}'' and, therefore, she should be able to explain the decision. Transparency is also seen as key for users to develop \emph{trust} toward the system~\cite{491,744,630}. In some studies that focus on transparency, trust is thus not the direct purpose of the system's explanation facility, but an expected indirect effect of transparency. In a number of other works, however, trust-building is explicitly mentioned as the goal of the explanations.

The second most frequent purpose of explanations in the analysed primary studies is \emph{effectiveness}, i.e.\ to help users assess if the recommended alternative is truly adequate for them.\footnote{One of the examined studies~\cite{423} focused on \emph{safety} in the context of decisions with critical consequences. We included it in the category \emph{effectiveness}.} \emph{Persuasiveness}, i.e.\ a system's capability to nudge the user to a certain direction, which can be conflicting with effectiveness~\cite{606}, was also in the focus of a number of studies.

Looking at the historical developments, we can observe in Figure~\ref{fig:purpose} that other potential purposes, like user satisfaction, scrutability, and efficiency, received more attention in the recent past. Reducing the user's cognitive load and trying to increase their satisfaction with the system (thereby increasing their intention to return) are essential aspects for e-commerce applications, which have been increasingly investigated during the past few years. In this commercial context, the potential persuasive nature of explanations~\cite{Bilgic:IUI-BP2005:ExplainingRecommendations} also attracted more research interest recent years.


\subsubsection{Underlying Decision Inference Methods}

Given that in most cases the explanation generation process is highly coupled with the underlying inference method, we analysed which methods are used in the investigated primary studies to infer the suggested alternative. Table~\ref{tbl:DMalgorithm} shows the outcomes of this analysis, in which one study may fall into more than one category, if it uses different methods.

\begin{table}
	\scriptsize
	\centering
	\caption{Decision Inference Methods: Method Type by Decade.}
	\label{tbl:DMalgorithm}
	\renewcommand{\tabcolsep}{0.55mm}
	\begin{tabular}{l l r r r r r r r l}
			\toprule
			\textbf{Category} & \textbf{Subcategory} & \textbf{1980} & \textbf{1990} & \textbf{2000} & \textbf{2010} & \textbf{Total} & \textbf{\%} \\ \midrule
				Knowledge-based & & 33 & 50 & 28 & 31 & 142 & 68.3\% \\
				\midrule
				 & \emph{Rule-based} & \emph{28} & \emph{33} & \emph{11} & \emph{7} & \emph{79} & \emph{55.6\%} \\
				 & \emph{Logic-based} & \emph{2} & \emph{3} & \emph{3} & \emph{8} & \emph{16} & \emph{11.3\%} \\
				 & \emph{Multi-Criteria Decision Making} & \emph{0} & \emph{1} & \emph{3} & \emph{5} & \emph{9} & \emph{6.3\%} \\
				 & \emph{Constraint-based} & \emph{0} & \emph{1} & \emph{1} & \emph{0} & \emph{2} & \emph{1.4\%} \\
				 & \emph{Case-based Reasoning} & \emph{0} & \emph{1} & \emph{3} & \emph{2} & \emph{6} & \emph{2.9\%} \\
				 & \emph{Other} & \emph{3} & \emph{11} & \emph{7} & \emph{9} & \emph{30} & \emph{21.1\%} \\  \midrule
				Machine Learning & & 2 & 13 & 12 & 24 & 51 & 24.5\% \\
				\midrule
				 & \emph{Feature-based} & \emph{2} & \emph{13} & \emph{6} & \emph{11} & \emph{32} & \emph{62.7\%} \\
				 & \emph{Collaborative-filtering} & \emph{0} & \emph{0} & \emph{5} & \emph{5} & \emph{10} & \emph{19.6\%} \\
				 & \emph{Hybrid} & \emph{0} & \emph{0} & \emph{1} & \emph{8} & \emph{9} & \emph{17.6\%} \\
				\midrule
				Mathematical Model & & 0 & 2 & 0 & 0 & 2 & 1.0\% \\
				\midrule
				Human-made Decision & & 0 & 1 & 0 & 1 & 2 & 1.0\% \\
				\midrule
				Algorithm-independent & & 0 & 3 & 3 & 5 & 11 & 5.3\% \\
				\bottomrule
	\end{tabular}
\end{table}

Given the historical importance of explanations in the context of (rule-based) expert systems, it is not surprising that the majority of the examined studies adopted a knowledge-based approach for decision inference and, correspondingly, for generating the explanations. Again, we can see the declining role of rule-based systems over the years and an increasing adoption of approaches based on ML. We use a two-level categorisation scheme to be able to detect the developments over time in a fine-grained manner. Based on such a categorisation, we can see, for example, that explanations for collaborative-filtering recommender systems are only investigated after the year 2000.

The subcategory \emph{Other} covers various alternative forms of knowledge-based approaches, including those that use special heuristics or ontologies, as well as studies that state that they use a form of knowledge-based reasoning without providing further details. Besides two studies that use mathematical models to infer the suggested alternative, there are two approaches~\cite{532,845} in which a decision is actually provided by the user, and the system seeks to complement this decision with an explanation. Their goals are: (i) to record the decision rationale for future inspections; or (ii) to provide an explanation on behalf of decision makers, who have expertise on the domain, to save their time.

Finally, Table~\ref{tbl:DMalgorithm} shows that only 11 approaches discuss explanation generation approaches that are independent of the underlying inference method. In some sense, this is not surprising, because generating an explanation using solely the user inputs or context together with the selected alternative is not trivial. However, this deserves further investigation, because extracting explanations from today's decision inference methods is becoming increasingly complex due to the increasing complexity of widely adopted ML algorithms, which may be even confidential, as argued by both Zanker and Ninaus~\cite{428} and Vig et al.~\cite{386}. These authors decouple explanations from the underlying inference method by proposing the so called \emph{knowledgeable explanations} and \emph{tag-based explanations}, respectively. This can lead to explanations that may be disconnected from the reasons of why a decision inference method suggested a particular alternative. Although this alleviates the problems discussed above, it compromises system transparency. Another direction is the use of ML algorithms that allow to produce \emph{explainable recommendations}, proposed by Zhang~\cite{596}, which use latent factor models.

\subsection{\ref{i:rq3} \RQEVAL}

We now focus on \emph{how} explanation generation approaches were \emph{evaluated} and, in some cases, compared. We first discuss whether the respective proposed forms of explanations were evaluated in the paper in which they were published. We then detail the types of evaluations adopted as well as the domains chosen to perform them. Finally, as the most common way of evaluating explanations is by means of user studies, we report their characteristics, such as their independent and dependent variables as well as sample size.

\subsubsection{Presence of an Evaluation}

Considering \textsc{technique} and \textsc{tool} studies, we investigated to what extent explanations were evaluated in the paper that they were proposed. In this analysis, we considered any form of evaluation except cases in which the authors simply described a simple scenario (i.e.\ a \emph{toy example}) to illustrate the use of the proposed approach and the explanation produced, even if this toy example was referred to as case study by the authors.

The results are shown in Table~\ref{tbl:approachToolEvaluation}, in which we can observe that less than a quarter (21.5\%) of the studies involving \textsc{techniques} and \textsc{tools} contains any form of evaluation, apart from toy examples. This can be partially explained by the fact that, when many expert systems papers were published (in the 1980s and 1990s), the methodological requirements in this field were probably lower in terms of evaluations than they are today.

\begin{table}
	\scriptsize
	\centering
	\caption{\textsc{Technique} and \textsc{Tool} Studies with Evaluation (Number and Percentage).}
	\label{tbl:approachToolEvaluation}
	\begin{tabular}{l r r r r r r r r r r}
			\toprule
			 & \multicolumn{2}{c}{\textbf{1980}} & \multicolumn{2}{c}{\textbf{1990}} & \multicolumn{2}{c}{\textbf{2000}} & \multicolumn{2}{c}{\textbf{2010}} & \multicolumn{2}{c}{\textbf{Total}} \\
			 & \textbf{\#} & \textbf{\%} & \textbf{\#} & \textbf{\%} & \textbf{\#} & \textbf{\%} & \textbf{\#} & \textbf{\%} & \textbf{\#} & \textbf{\%} \\
			\midrule
			\textsc{Technique} & 0 & 0.0\% & 5 & 21.7\% & 11 & 50.0\% & 19 & 46.3\% & 35 & 34.7\% \\
			\textsc{Tool} & 0 & 0.0\% & 3 & 7.7\% & 1 & 5.6\% & 1 & 6.7\% & 5 & 5.6\% \\
			\midrule
			\textbf{Total} & 0 & 0.0\% & 8 & 12.9\% & 12 & 30.0\% & 20 & 35.7\% & 40 & 21.1\% \\
			\bottomrule
	\end{tabular}
\end{table}%

Nevertheless, it is surprising that even nowadays (from 2010-present), the presence of an evaluation accompanying the proposal of a new form of explanation is still not typical, with almost two thirds of all analysed studies lacking a proper evaluation. In some studies, this can be explained by the fact that the main focus of the work was on another contribution, e.g.\ a recommendation algorithm or an expert system that included an explanation facility, and the evaluation is then limited to this main contribution.

\subsubsection{Evaluation Types and Domains}

\begin{figure}
	 \centering
   \includegraphics[width=\linewidth]{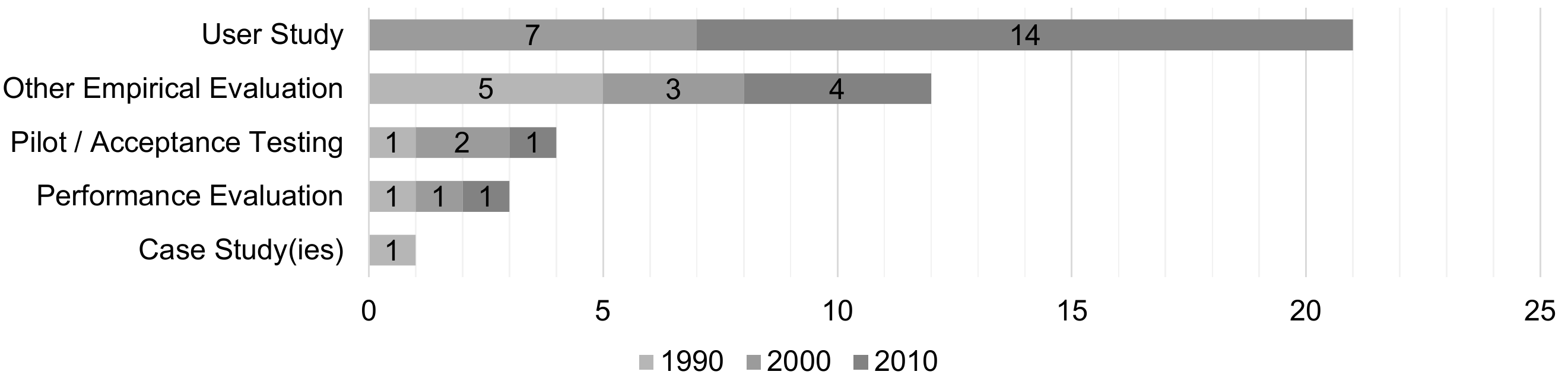}
   \caption{Evaluation Types.}
   \label{fig:evalTypes}%
\end{figure}

Next, we analyse which \emph{type} of evaluation researchers applied to assess or compare different explanations provided by a system. Figure~\ref{fig:evalTypes} shows the five main types of evaluation that we found in the primary studies in which new forms of explanations are proposed. As can be seen, all identified evaluation types are empirical. We do not include the 22 \textsc{evaluation} studies in Figure~\ref{fig:evalTypes}, because all of them describe results of user studies.

User studies are the predominant research method used to evaluate proposed \textsc{techniques} and  \textsc{tools}, occurring in more than half (52.4\%) of the cases. This is expected because there is no formal definition of a \emph{correct} or \emph{best} explanation in many application scenarios. In these cases, the only way to evaluate the provided explanations is to capture the subjective perception of the users or to monitor the impact of the explanations in the user behaviour.

Possibly due to the time required to conduct user studies, alternative evaluation types, which do not require the availability of participants, were the choice in the remaining studies. Most of these, 12 in total, included a customised form of empirical evaluation, which involved generating explanations based on the proposed approach, and collecting measurements that are possibly specific to the explanation problem. An example of such a measurement is \emph{explanation coverage}~\cite{1028}, which is defined as the fraction of features that are part of the user preferences that are used in the explanation. Three studies focused specifically on the performance (computational efficiency) of their approach. Finally, the remaining evaluations either comprised a pilot study, or alternatively an acceptance test, or a set of case studies (reported in a single study~\cite{834}).

\begin{figure}[t]
	 \centering
     \subfloat[Evaluations per Domain.\label{fig:evalDomains}]{%
       \includegraphics[width=\linewidth]{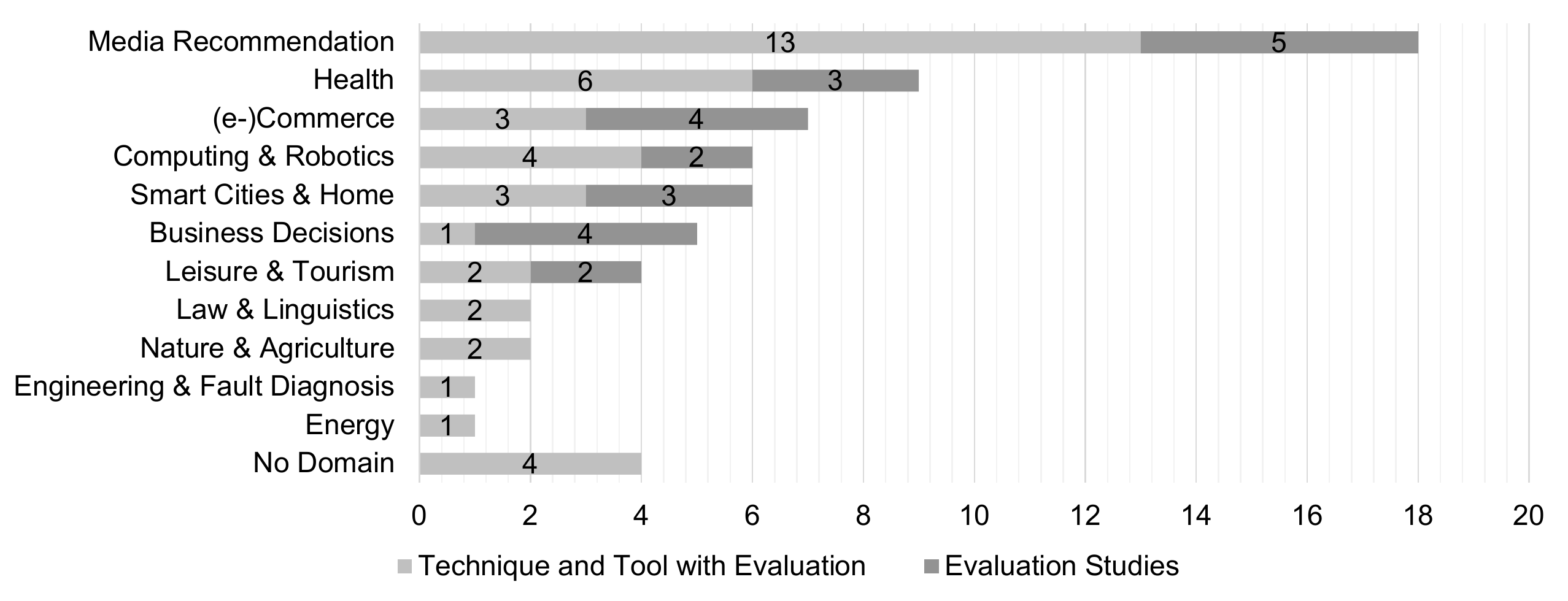}
     }
		\\
     \subfloat[Number of \textsc{Tool} Studies per Domain.\label{fig:toolDomain}]{%
       \includegraphics[width=\linewidth]{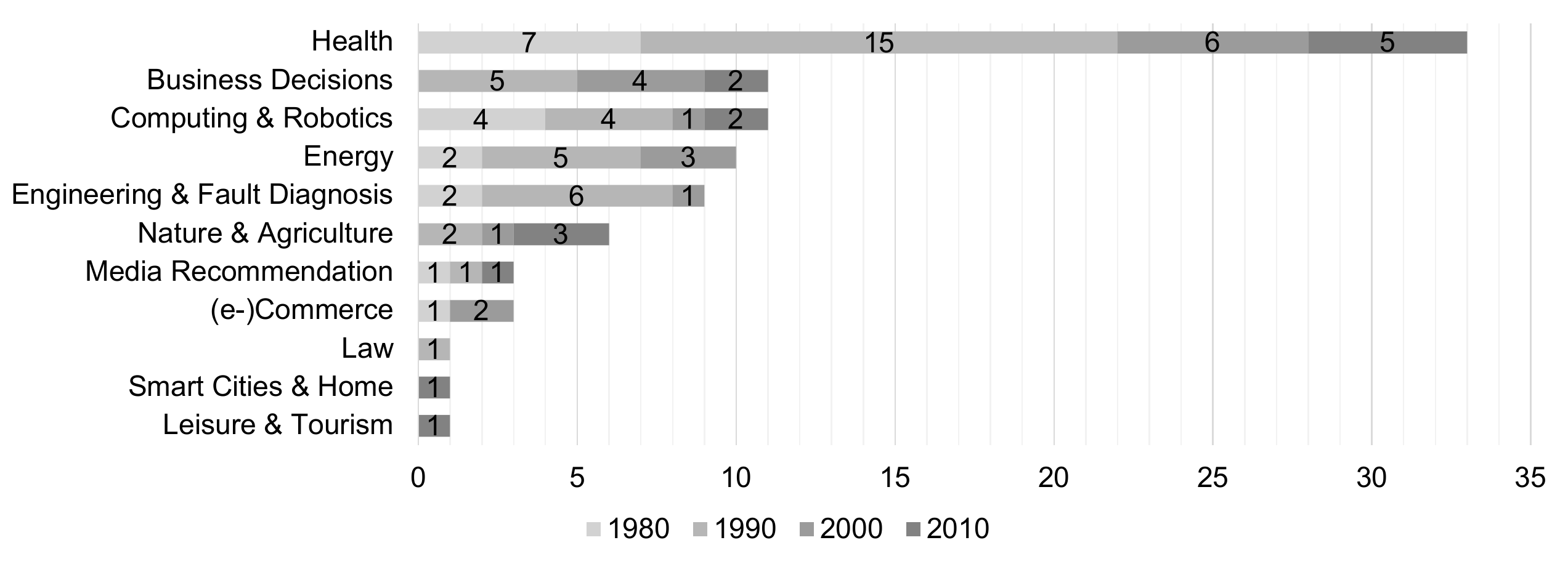}
     }
     \caption{Domain Analysis.}
     \label{fig:domainanalysis}
\end{figure}

Having looked at the evaluation type, we now analyse which domains researchers chose to perform their evaluations. Figure~\ref{fig:evalDomains} summarises the results of the analysis of all studies that include an evaluation. The results show that the most common application domain is \emph{Media Recommendation}, followed by \emph{Health}. In addition, a number of evaluations was done in the context of movie or music recommendation (which are included in the \emph{Media Recommendation} category) in the past few years. The same applies to the domain of \emph{(e-)Commerce}. In four studies, although there is an evaluation, the selected application domain is not reported (\emph{No Domain} in Figure~\ref{fig:evalDomains}).

To understand how the interest in particular application domains changed over time, we present in Figure~\ref{fig:toolDomain} the domains associated with studies that describe a \textsc{tool}. Such \textsc{tools} were more frequently proposed in the past decades, as opposed to evaluation studies, which received more attention in recent years. Considering \textsc{tools}, the \emph{Health} domain has been consistently in the focus of researchers over time. However, although it is the second most targeted domain in evaluations, the comparison between Figures~\ref{fig:evalDomains} and~\ref{fig:toolDomain} shows that the \emph{Health} domain has been largely more explored in the context of \textsc{tools} than in evaluations. The popularity of this domain in developed \textsc{tools} is mainly due to the many expert systems of the 1980s and 1990s that were designed to support medical decision making. Furthermore, we can observe that many domains were not explored in the past few years, but this is also a consequence of the generally declining number of studies that describe \textsc{tools}.

\subsubsection{User Studies}

We now investigate the user studies in more depth. We first discuss their design details in terms of the \emph{independent and dependent variables}, followed by an analysis of their \emph{sample size}, i.e.\ the number of involved participants.

\paragraph{Independent Variables}

\begin{figure}
	 \centering
   \includegraphics[width=\linewidth]{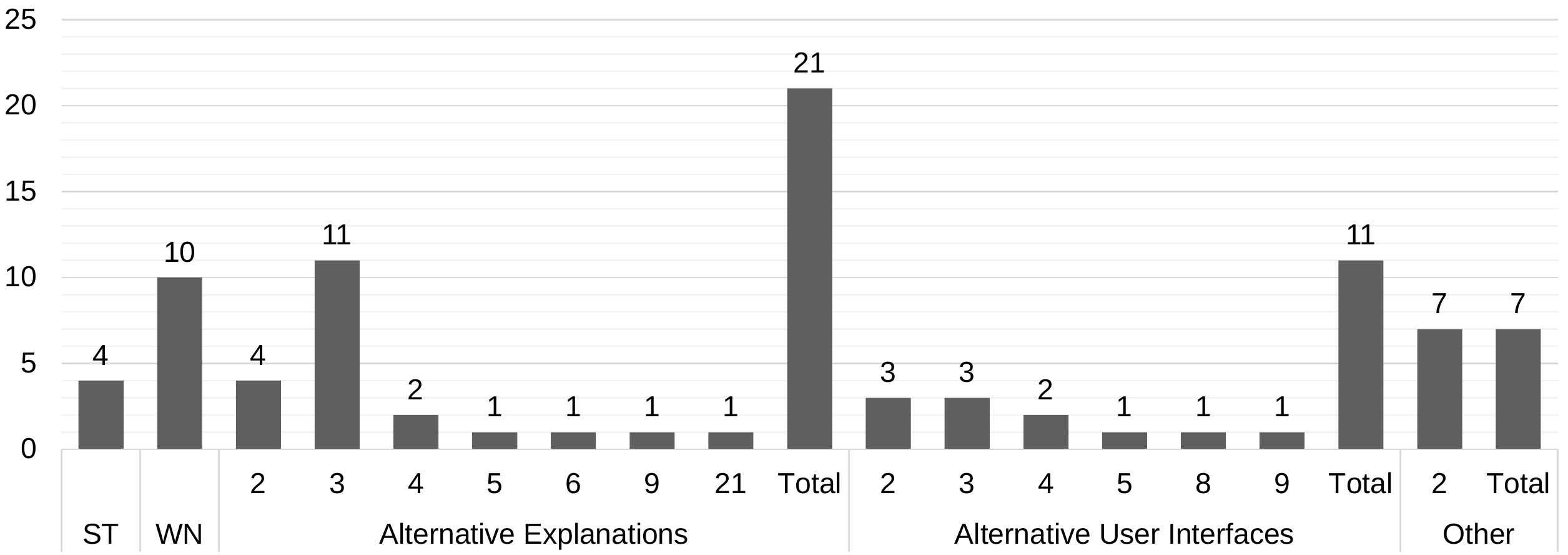}
   \caption{Types of Independent Variables used in User Studies.}
   \label{fig:evalTreatments}
\end{figure}

Figure~\ref{fig:evalTreatments} shows the independent variables of the study designs. In a few (four) cases, only one single treatment was used (marked as ST). In ten studies, explanations were presented in one condition but not in the other (marked as WN, standing for \emph{With explanations} and \emph{No explanations}). The majority of the studies, however, compared different kinds of explanations or the impact of their presence when providing a recommendation. In some of these studies, one of the alternatives was providing participants with no explanation. The label at the bottom of the bars in Figure~\ref{fig:evalTreatments} indicates how many alternatives were compared, while the label at the top provides the number of studies that had this number of alternatives.

We distinguished cases in which the study focused solely on different types of explanations (i.e.\ the only varying component of the information presented to the user is the explanation, while other components remain fixed) and cases in which the focus was on aspects of the user interface (i.e.\ presence or absence of alternative interface components, which influence the provided explanation). The former cases are referred to as \emph{alternative explanations} in Figure~\ref{fig:evalTreatments}, while the latter are considered \emph{alternative user interfaces}.

A smaller amount of studies (\emph{Other}, in Figure~\ref{fig:evalTreatments}) investigated other relevant explanation-related aspects, such as the impact of the user's expertise level and different explanation properties. Examples of such properties are: (i) \emph{source}, which indicates from where the information presented in the explanation is extracted; (ii) \emph{length}, measured by the number of characters or number of features used in an explanation; (iii) \emph{direction}, which can either be positive and aimed at justifying why a certain alternative fits a certain user, or negative, when it justifies the relaxation of certain input conditions; and (iv) \emph{confidence}, which is associated with the vocabulary used in explanations, indicating how confident the system is with respect to the suggestion made to the user.

\paragraph{Dependent Variables}

\begin{figure}
	 \centering
   \includegraphics[width=\linewidth]{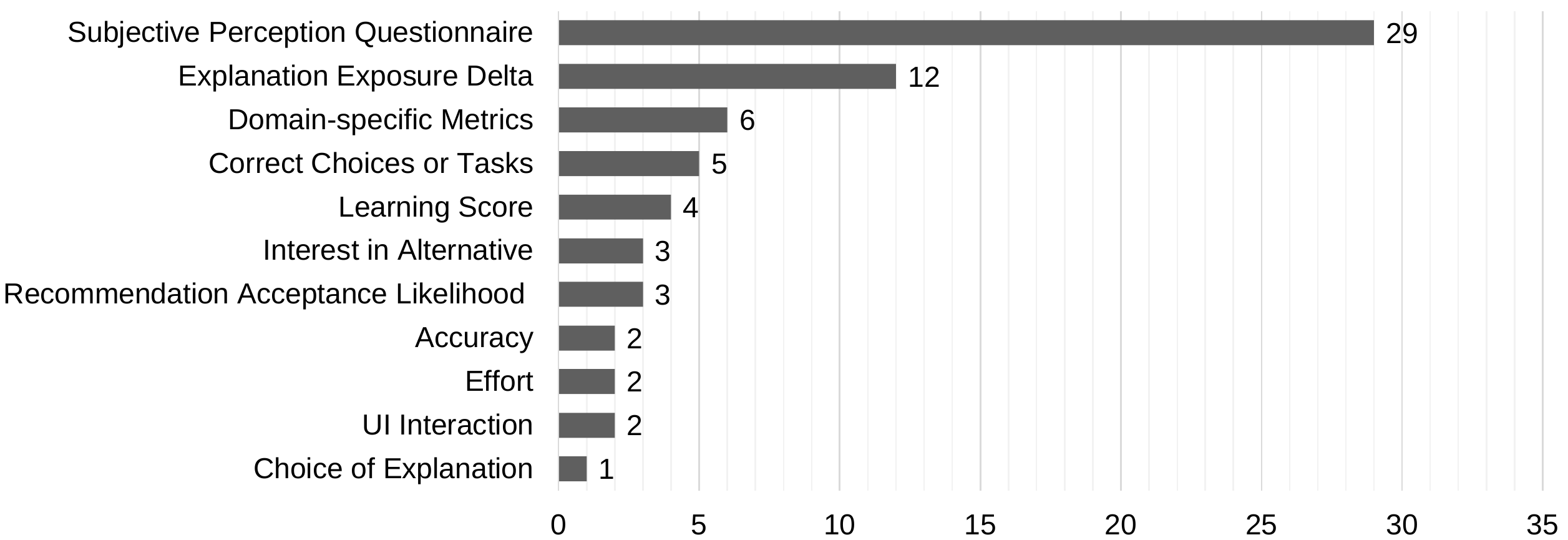}
   \caption{Types of Dependent Variables used in User Studies.}
   \label{fig:evalMeasurements}
\end{figure}

The different measurements that were collected in the user studies are summarised in Figure~\ref{fig:evalMeasurements}. Note that several measurements were made in the majority of the studies. Various studies included the collection of the opinion of study participants on certain aspects. The second most adopted measurement, which we refer to as \emph{explanation exposure delta}, was first adopted by Bilgic and Mooney~\cite{Bilgic:IUI-BP2005:ExplainingRecommendations}\footnote{Bilgic and Mooney's work is not included in this review because it was published in a workshop and not part of the databases searched in our review. We discuss the choice of databases and possible research limitations later in the paper.} and uses a specific protocol to evaluate explanations. These types of dependent variables are described in Table~\ref{tbl:measurementTypes}, together with the many other variable types shown in Figure~\ref{fig:evalMeasurements}.


\begin{table}[t]
	\scriptsize
	\centering
	\caption{Types of Dependent Variables in User Studies.}
	\label{tbl:measurementTypes}
	\renewcommand{\tabcolsep}{0.8mm}
	\begin{tabularx}{\linewidth}{p{2.8cm} X}
			\toprule
			\textbf{Measurement Type} & \textbf{Description} \\
			\midrule
			Subjective Perception Questionnaire & Participants are asked a set of questions in order to obtain participants' subjective view with respect to different explanation aspects. Responses are typically collected using a Likert-scale. \\ \hline
			Explanation Exposure Delta & Measures the difference between a score given by participants before and after the presentation of an explanation. \\ \hline
			Domain-specific{\par}Metrics & Measurement of metrics that are meaningful only in particular domains, e.g.\ percentage of forecasts that were adjusted (forecasting domain). \\ \hline
			Correct Choice or Tasks & Evaluates how many correct choices (or accomplished tasks) participants are able to make, when there is a notion of choice correctness assumed in the study (e.g.\ a disease diagnosed based on symptoms). \\ \hline
			Learning Score & Measures how much participants learn by interacting with the system. \\ \hline
			Interest in Alternative & Measures to what extent participants are interested in the recommendation, e.g.\ by rating the recommended alternative. \\ \hline
			Recommendation{\par}Acceptance Likelihood & Measures whether participants agree with a recommendation or predicted suitability score. \\ \hline
			Accuracy & Measures the difference between the predicted suitability of suggested alternatives and how participants evaluate them considering provided explanations. \\ \hline
			Effort & Measures how much effort (in time) participants spend making a decision or evaluating an explanation. \\ \hline
			UI Interaction & Measures how participants interact with the user interface, e.g.\ number of clicks or requests to more detailed explanations. \\ \hline
			Choice of Explanation & Asks participants which from a set of alternative explanations they prefer. \\
			\bottomrule
	\end{tabularx}
\end{table}%

\begin{figure}
	 \centering
   \includegraphics[width=0.6\linewidth]{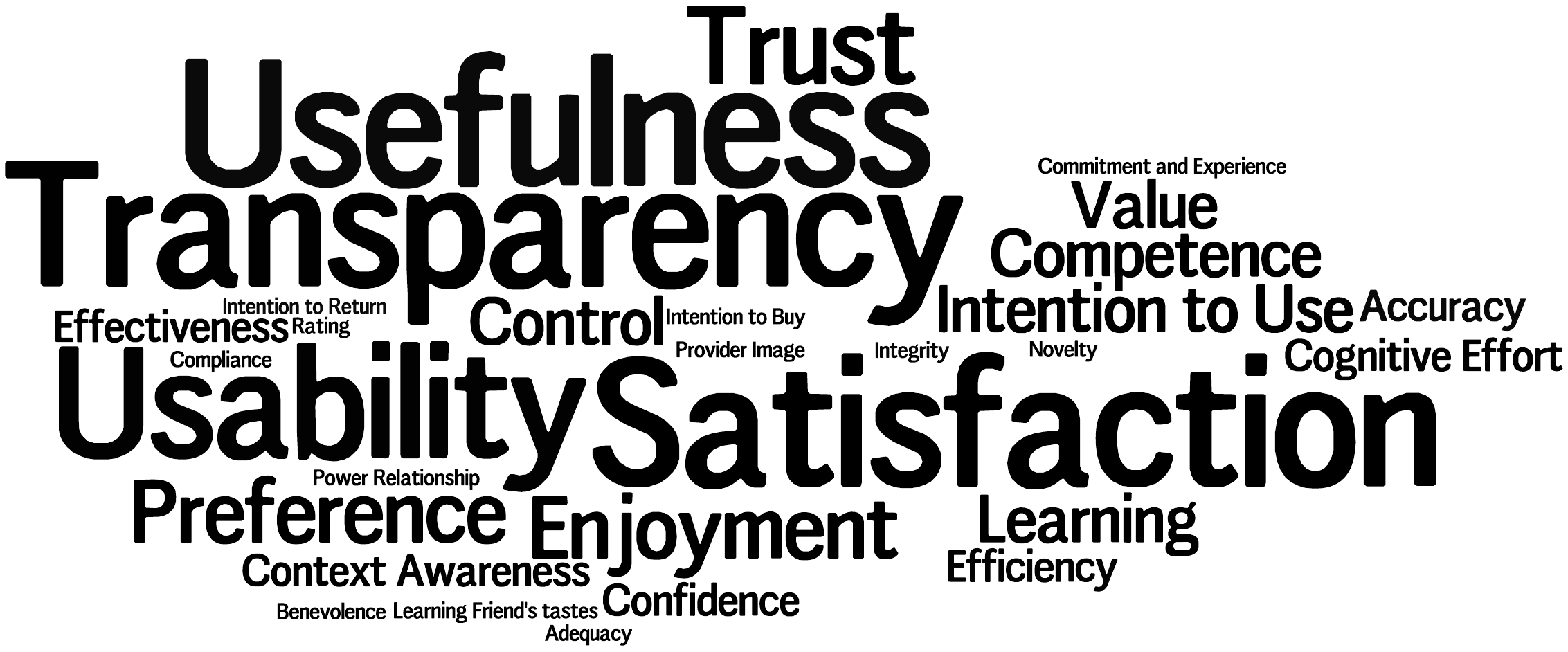}
   \caption{Tag Cloud of Question Topics in Subjective Perception Questionnaires.}
   \label{fig:questionnaireWordCloud}%
\end{figure}

Looking at the subjective perception questionnaires in more detail, we observed that participants were asked a wide variety of questions in the studies in order to investigate different aspects of explanations. We selected terms commonly used to refer to these aspects, given that there is no standardised terminology to classify the analysed aspects. Each question was classified using our selected terms. An example is asking the participant to indicate the agreement with the sentence: ``The user interface is easy to use,'' which is associated with \emph{usability}. Since such a classification approach leaves room for interpretation, we do not report specific occurrence numbers here, but represent the information in the form of a tag cloud (Figure~\ref{fig:questionnaireWordCloud}). The tag cloud shows that \emph{transparency} is one of the main aspects that are evaluated in user studies. This observation matches the results from Section~\ref{sec:purposes} (Figure~\ref{fig:purpose}), in which transparency is mentioned as the most frequently stated intended purpose of providing explanations. A similar observation can be made for \emph{trust}. Interestingly, \emph{satisfaction} was often in the focus of the questionnaires, even though it is not commonly listed as an investigated explanation purpose. Note that satisfaction, according to Tintarev and Masthoff~\cite{296}, refers to \emph{usefulness} and \emph{usability}, but they are often explicit targets of specific questions.

\paragraph{Sample Size}

\begin{figure}
	 \centering
   \includegraphics[width=0.7\linewidth]{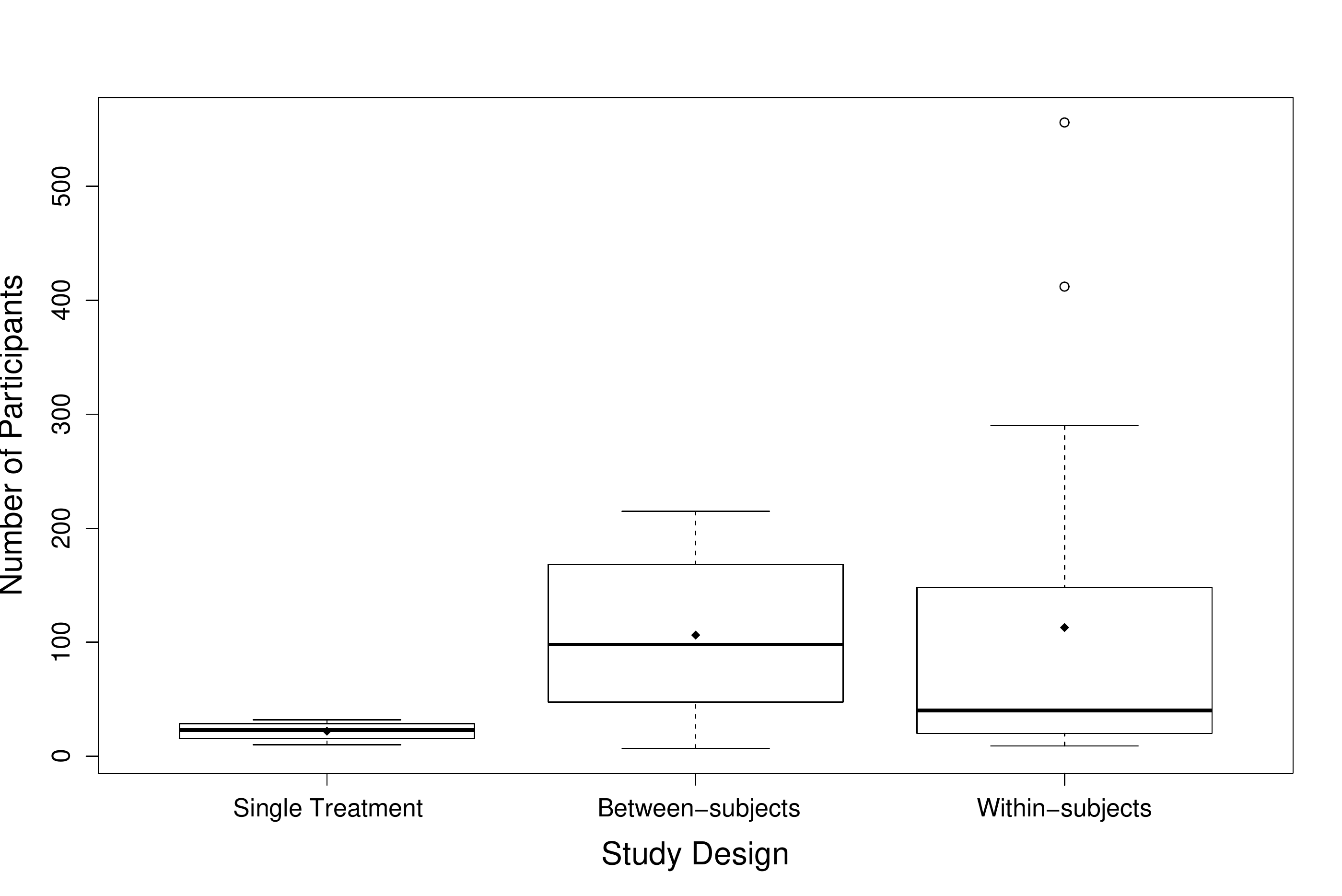}
   \caption{Sample Size in User Studies.}
   \label{fig:participants}%
\end{figure}

We now consider the number of subjects that were involved in the user studies. We differentiate between three study designs: (i) single treatment, in which no comparison is made; (ii) between-subjects, in which subjects are split into groups and receive different treatments; and (iii) within-subjects, in which all subjects experience all of the different treatments. Typically, between-subjects studies should have a higher number of participants to achieve statistical significance. We present a box plot of the number of participants according to the different study designs in Figure~\ref{fig:participants} and provide the exact numbers and further details in Table~\ref{tbl:participants}. Two between-subjects studies were not considered in this analysis because they do not report the number of participants. Moreover, when there is more than one user study reported in the same work, we considered each user study individually.

\begin{table}[t]
	\scriptsize
	\centering
	\caption{Number of Participants in User Studies.}
	\label{tbl:participants}
	\begin{tabular}{l r r r r r r r r r r r}
			\toprule
			\textbf{Study Design} & \textbf{Number of}  & \textbf{Mean}  & \textbf{Standard} & \textbf{Median}  & \textbf{Min} & \textbf{Max} \\
			                      & \textbf{Studies}    &                & \textbf{Deviation} &                 &                  &                  \\
			\midrule
			Single Treatment     & 4                    & 22.0           & 9.2                         & 23             & 10             & 32             \\
			Between-subjects     & 20                   & 106.3          & 68.4                        & 98             & 7              & 215            \\
			Within-subjects      & 21                   & 112.9          & 147.8                       & 40             & 9              & 556            \\
			\midrule
			\textbf{All Studies} & \textbf{45}          & \textbf{101.9} & \textbf{112.2}              & \textbf{54}    & \textbf{7}     & \textbf{556}   \\
			\bottomrule
	\end{tabular}
\end{table}%

Even though the single largest study had a within-subjects design, between-subjects studies indeed had the largest median of participants. Moreover, not only the study design should influence the number of required participants, but also the number of factors that are investigated. However, we did not observe a relationship between these aspects. For example, the study performed by Guy et al.~\cite{437} with 412 participants is a within-subjects study and has five treatments (different versions of recommender systems). In contrast, Felfernig and Gula~\cite{272} performed a study with 116 participants followed a between-subjects design and evaluated eight different versions.
Finally, we observe that the number of participants of single treatment studies is typically low. A possible reason is that empirical studies are expected in an increasing number of subfields of computer science, and researchers resort to comparably simple studies in order to provide some form of evaluation. In any case, the low number of participants involved in these studies is a potential threat to their external validity.

\subsection{\ref{i:rq4} \RQCONC}

After analysing how researchers evaluate different forms of explanation, we now focus on the conclusions they reached. This analysis also includes the main findings of the few \textsc{foundational} studies that we identified.

\subsubsection{Analysis Method}

A number of primary studies associated with \textsc{techniques} and \textsc{tools} only provided anecdotal evidence that the proposed approach is generally feasible or produces reasonable explanations. Additionally, in some cases, data was gathered to describe general characteristics of the generated explanations, such as their length. Such forms of unstructured evaluation, unfortunately, only provide limited evidence of the true benefits of the proposed approaches. We next thus only analyse conclusions that are based on user studies, including those published in \textsc{evaluation} papers.

As there is no standard design for user studies that evaluate forms of explanations, combining the results reported in the literature is not trivial. Again, to avoid researcher bias, we devised the following systematic way to analyse and contrast the obtained results. First, we extracted explicitly stated \emph{key observations} and \emph{conclusions} that were made by the authors of the studies. In this step, we did not infer any additional conclusions based on the provided data and also did not question the validity of the conclusions stated by the authors. Second, we classified the conclusions according to the explanation characteristics that were investigated in the studies. Often, these characteristics match possible explanation purposes and, in case they were not explicitly stated, we inferred this information from the measurements.\footnote{Some authors measured the \emph{accuracy} of the decisions or ratings of the study participants. Such a measurement is categorised as investigating the \emph{effectiveness} of the explanations.} Third, we organised the conclusions as tuples of the form $\langle$\texttt{direction target-explanation-style} [\texttt{compared-explanation-styles}]$\rangle$. The direction can be positive, negative, or neutral, indicating that the target (proposed) explanation style increases, decreases, or has no impact in a certain measurement when compared with other explanation styles. We use the term \emph{explanation style} to refer to the specific forms of explanation considered in the studies. When no information about other explanation styles is given, it means that the baseline is the provision of no explanations.


\subsubsection{Conclusions Reached in User Studies}

We split the results of our analysis, i.e.\ the set of recorded tuples, into three parts. Studies that found positive effects of providing explanations \emph{vs.\ }providing no explanations are listed in Table~\ref{tbl:positiveResultsNo}. Studies that report positive effects in a comparison with alternative explanation styles are given in Table~\ref{tbl:positiveResultsOther}. Studies with neutral and negative conclusions are shown in Table~\ref{tbl:neutralNegativeResults}.

\begin{table}[t]
	\scriptsize
	\centering
	\caption{User Studies with Positive Results with respect to No Explanations.}
	\label{tbl:positiveResultsNo}
	\begin{tabular}{l l}
			\toprule
			\textbf{Purpose} & \textbf{Explanation Style} \\ \midrule
			Effectiveness & Confidence+Sensor Data with Low Robot Ability~\cite{679} \\
										& Decisive Input Values~\cite{817} \\
			              & Derived Topic Models and Time Intervals~\cite{633} \\
			              & Explanation Interfaces~\cite{1017} \\
                    & Peer Graph Navigation~\cite{356} \\
                    & Tag-based~\cite{386} \\	
			\hline
			Transparency & Confidence+Sensor Data with Low Robot Ability~\cite{679} \\
					         & Decisive Input Values~\cite{817} \\
									 & Derived Topic Models and Time Intervals~\cite{633} \\
			             & Justification~\cite{1058} \\
			             & POMDP* Translation~\cite{678} \\
                   & Tag-based~\cite{386} \\
			\hline
			Persuasiveness & Peer Information~\cite{381} \\
                     & Peer- and Tag-based~\cite{437} \\
			\hline
			Satisfaction & Justification~\cite{1058} \\
									 & Music Domain Knowledge~\cite{1188} \\
                   & Pros and Cons~\cite{272} \\
			\hline
			Trust & Confidence+Sensor Data with Low Robot Ability~\cite{679} \\
						& How/Why/Trade-off~\cite{1183} \\
			      & POMDP* Translation~\cite{678} \\
            & Pros and Cons~\cite{272} \\
			\hline
			Usefulness & Decisive Features~\cite{655}	\\	
                 & Match Score+Decisive Features~\cite{428} \\
                 & Match Score+Decisive Features~\cite{530}	\\					
			\hline
			Ease of Use & Explanation Interfaces~\cite{1017} \\
			\hline
			Efficiency & Decisive Features~\cite{428} \\
			\hline
			Education & Decisive Features~\cite{455} \\
			\bottomrule
			\multicolumn{2}{l}{* POMDP stands for partially observable Markov decision process.}
	\end{tabular}
\end{table}%

\begin{table}[t]
	\scriptsize
	\centering
	\caption{User Studies with Positive Results with respect to Alternative Explanations.}
	\label{tbl:positiveResultsOther}
	\begin{tabular}{l l}
			\toprule
			\textbf{Purpose} & \textbf{Explanation Style} \\ \midrule
			Effectiveness & Non-personalised Decisive Features (Personalised, Popularity): Cameras~\cite{1076} \\
			              & (Non-)personalised Tag-based (Keyword-based)~\cite{491} \\
										&	(Non-)personalised Tag-based (Popularity, Neighbours)~\cite{1131} \\
                    & Predicted Rating (Popularity, Single Feature)~\cite{1173} \\
			              & Trace (Justification, Strategy)~\cite{1073} \\	
                    & Visualisation (Histogram)~\cite{511} \\	
			\hline
			Transparency & Personalised Tag-based (Non-personalised, Confidence, Neighbours)~\cite{1131} \\
                   & Visualisation (Histogram)~\cite{511} \\
			\hline
			Persuasiveness & Histogram (20 explanations)~\cite{208} \\
			               & Justification (Trace, Strategy)~\cite{861} \\
                     & Personalised Features (People also Viewed, No explanation)~\cite{596} \\
                     & Social Explanations (Peers, Personalisation)~\cite{550} \\
                     & Social Explanations+Decisive Features (Authority, Social Proof)~\cite{598} \\						 
			\hline
			Satisfaction & Personalised Decisive Features (Non-personalised, Popularity): Movies I~\cite{1076} \\
			             & Personalised Decisive Features (Non-personalised): Cameras~\cite{1076} \\
								 	 & Non-personalised Decisive Features (Popularity): Movies Final~\cite{1076}  \\
									 & (Non-)personalised Tag-based (Keyword-based)~\cite{491}  \\
									 & Personalised Tag-based (Popularity, Neighbours)~\cite{1131} \\
                   & Predicted Rating (Popularity, Single Feature)~\cite{1173} \\
			\hline
			Trust & Confidence+Decisive Features (Confidence)~\cite{651} \\
			\hline
			Usefulness & Contextualised Deep Explanations (Non-contextualised)~\cite{943} \\
							   & Trace (Justification, Strategy)~\cite{1073} \\					
			\hline
			Ease of Use & Contextualised Deep Explanations (Non-contextualised)~\cite{943}  \\
                  & Tag-based (Item-based, Feature-based)~\cite{566} \\
			\hline
			Efficiency & (Non-)personalised Tag-based (Keyword-based)~\cite{491} \\
			           & (Non-)personalised Tag-based (8 explanations)~\cite{1131}\\
			\hline
			Education & \\
			\bottomrule
	\end{tabular}
\end{table}%

\begin{table}
	\scriptsize
	\centering
	\caption{User Studies with Neutral or Negative Results.}
	\label{tbl:neutralNegativeResults}
	\begin{tabularx}{\linewidth}{l X X}
			\toprule
			\textbf{Purpose} & \multicolumn{2}{c}{\textbf{Explanation Styles with}} \\ \cline{2-3}
			& \textbf{Neutral Results} & \textbf{Negative Results}  \\\midrule
			Effectiveness & Confidence+Sensor Data with High Robot Ability~\cite{679}{\par}
			                Hierarchic+Deep Explanations~\cite{955}{\par}
										  Match Score (No recommendation, Recommendation)~\cite{487}{\par}
                      MovieLens~\cite{208}{\par}
											(Non-)personalised Decisive Features (Popularity): Movies I~\cite{1076}{\par}
											Non-personalised Decisive Features (Personalised): Movies II~\cite{1076}
											& (Non-)personalised Decisive Features (Popularity): Movies Final~\cite{1076}\\
			\hline
			Transparency & Confidence+Sensor Data with High Robot Ability~\cite{679} &  \\
			\hline
			Persuasiveness &   & Tag-based~\cite{437} \\
			\hline
			Satisfaction & Personalised Decisive Features (Non-personalised): Movies II~\cite{1076} \\
									 & Trace (Justification, Strategy)~\cite{1073} & \\
			\hline
			Trust & Confidence+Sensor Data with High Robot Ability~\cite{679} & \\
			      & Match Score+Decisive Features~\cite{530} &  \\
			\hline
			Usefulness &  &  \\
			\hline
			Ease of Use & Match Score+Decisive Features~\cite{428} &  \\
                  & Match Score+Decisive Features~\cite{530} &	\\
			\hline
			Efficiency &  & Derived Topic Models and Time Intervals~\cite{633} \\
			           &  & Pros and Cons~\cite{272}   \\
			\hline
			Education & Domain Knowledge~\cite{789} &  \\
			\bottomrule
	\end{tabularx}
\end{table}%

The largest amount of conclusions reached in the studies are related to the \emph{effectiveness} of explanations, typically measured by the \emph{explanation exposure delta}, which in this case the lower, the better.\footnote{Higher deltas mean that users tend to overestimate or underestimate suggested alternatives based on explanations.} These conclusions are also those that diverge the most. Given that the reported results concern different explanation styles, the observed divergence means that specific forms of explanations lead to more effective decisions. Moreover, this also suggests that there may be confounding variables in some studies, such as the accuracy of the underlying decision inference method and the study domain, which may influence the observed outcomes. Three studies illustrate the possible existence of such confounding effects. A study in the robotics domain~\cite{679} showed that explanations lead to higher effectiveness only \emph{when the robot ability is low}. Ehrlich et al.~\cite{487}, who initially observed no statistical difference in their user study, based on a finer-grained analysis of their results, concluded that explanations are helpful \emph{when the correct recommendation is provided}, which is not the case in the absence of such a recommendation. Furthermore, the four user studies reported by Tintarev and Masthoff~\cite{1076}, which involved more than one domain, led to slightly different results regarding effectiveness---some results were not significant while others provided evidence that presenting popularity-based or non-personalised decisive features are more effective than presenting decisive features in a personalised way.

Contradicting results were also observed when the goal of the studies included the investigation of the \emph{persuasiveness} of explanations. The data in Table~\ref{tbl:positiveResultsNo} shows that persuasiveness was mainly achieved when explanations are based on social information, such as peer ratings. Negative results were obtained when tags were used as a basis for explanations. In addition, the studies that compare different explanation styles (Table~\ref{tbl:positiveResultsOther}) confirm the value of social information when designing persuasive explanations~\cite{208,550,598}. Only in one single study~\cite{596}, it turned out that a traditional explanation of the form ``people also viewed'' was less persuasive than \emph{personalised features}. These are decisive features of an alternative selected based on preferences of the user receiving the recommendation.

\emph{Transparency} and many of the user-centric purposes---\emph{trust}, \emph{satisfaction}, and \emph{usefulness}---share similar results. Most of the studies indicate that explanations can in fact help to achieve these purposes. If not, the results do not provide evidence of negative effects.
This is not the case, however, of \emph{ease of use}. There are two studies associated with no effect in that respect, and only one reporting improvement. This improvement was achieved not only through the provision of explanatory information to users, but an enhanced user interface that categorises the suggested alternatives, possibly helping the user while analysing them. The non-existence of an effect on ease of use in the other studies is probably caused by the increased cognitive load for the users when more explanatory information is displayed. The fact that users have more information to process in such situations also explains the mostly negative effects of the provision of explanations on \emph{efficiency}.

With respect to the purpose of \emph{education}, there are only two studies, which reached different conclusions. Furthermore, none of the analysed user studies focused on the remaining purposes of explanations mentioned in the literature, namely \emph{scrutability} and \emph{debugging}.

Aside from the explanation purposes, some studies analysed the orthogonal aspect of \emph{personalising} explanations. Gedikli et al.~\cite{1131} showed that a personalised version of an explanation approach based on tag clouds led to higher levels user-perceived transparency than the non-personalised version and had a modest positive effect on satisfaction. However, the opposite can be observed with respect to effectiveness and efficiency. Their earlier study~\cite{491} indicates that personalisation had only a modest effect on efficiency, satisfaction, and effectiveness. Similarly, based on their four user studies, Tintarev and Masthoff~\cite{1076} concluded that personalised decisive features led to (sometimes modest) increased satisfaction. Nevertheless, this type of explanations caused a significant lower effectiveness in one of the studies.

A few studies are not reported in the summary tables as they focus on specific aspects of explanations. Ramberg~\cite{872} analysed the impact of \emph{different expertise levels} of the users and concluded that experts and novices have different preferences regarding the provided explanations. One study~\cite{860} investigated two explanation aspects: \emph{direction}, which can be positive or negative, and \emph{source}, which can be a decision support system or self-generated (by participants). The authors concluded that negative explanations are more influential than positive explanations, when they are generated by a decision support system. Finally, G\"{o}n\"{u}l et al.~\cite{998} focused on the particular explanation characteristics \emph{confidence} and \emph{length} and their study indicates that strongly confident and long explanations are more persuasive.

\subsubsection{Conclusions Reached in \textsc{Foundational} Studies}

Our literature search returned only a few \textsc{foundational} studies. A study performed by Tintarev and Masthoff~\cite{303} confirms the potential value of personalisation as discussed above. According to their findings~\cite{303}, explanations should be customised to the user, focusing on an appropriate set of features of the suggested alternative. They also suggest that explanations should be tailored to the context. Furthermore, they proposed two additional guidelines that state that (i) from the many features of alternatives, considering only a short list from which personalised features are selected is enough to satisfy most of the users; and (ii) the source of the explanations matters (e.g.\ peers mentioned in the explanation). Additional guidelines and patterns that complement this study were proposed by Nunes et al.~\cite{517} which, for example, state that explanations should be concise and focus on the most relevant criteria.

Three additional user studies did not evaluate aspects of the explanations themselves (thus not classified as \textsc{evaluation} studies), but investigated extrinsic aspects that contribute to a better understanding of explanations. The impact of \emph{mental models} was investigated by Rook and Donnell~\cite{823}, who concluded that it is important that users understand the expert system's reasoning process and the information provided in explanations to make good decisions. Giboney et al.~\cite{1167} found out that justifications that match user preferences (\emph{cognitive fit}) are valuable for increasing the acceptance of recommendations of knowledge-based systems. The results show that when this match occurs, users tend to be more engaged, leading to longer interaction times with the system. Finally, Gregor~\cite{936} investigated the usefulness of explanations in different contexts. The main outcome of the study is that explanations are more often accessed and helpful in cooperative problem solving situations. Moreover, when explanations are more often accessed, the problem-solving performance increases, particularly when system-user collaboration is required. Note that both these latter studies may indicate that more effective explanations are also those that lead to decreased efficiency.

\section{Discussion}\label{sec:sysRevDiscussion}

Based on the results from our systematic review, we present our insights in the field of explanations. Specifically, we (i) propose a new comprehensive taxonomy that captures the many facets that one might consider when designing an explanation facility for an \ags{}; and (ii) outline possible directions for future works. We also discuss limitations of our review.

\subsection{Explanation Taxonomy} 

A number of explanation taxonomies of different granularity levels has been proposed in the literature in the past~\cite{Ye:MISQ1995:ExplanationImpact,747,Gregor:MISQ1999:ExplanationSurvey,Lacave:KER2002:ExplanationSurveyBN,Lacave:KER2004:ExplanationSurvey,Nakatsu:chapter2006:ExplanationSurvey,1065,386,536,Friedrich:AIMag2011:ExplanationTaxonomy,694}. These taxonomies cover a variety of different aspects of explanation facilities, such as their purpose, the knowledge they use internally, or the information they convey to the user. Our review, however, revealed that there are a number of facets should be considered when designing a new explanation approach, which are not covered by these existing taxonomies.

\begin{figure}[p]
	\centering
	\includegraphics[width=\linewidth]{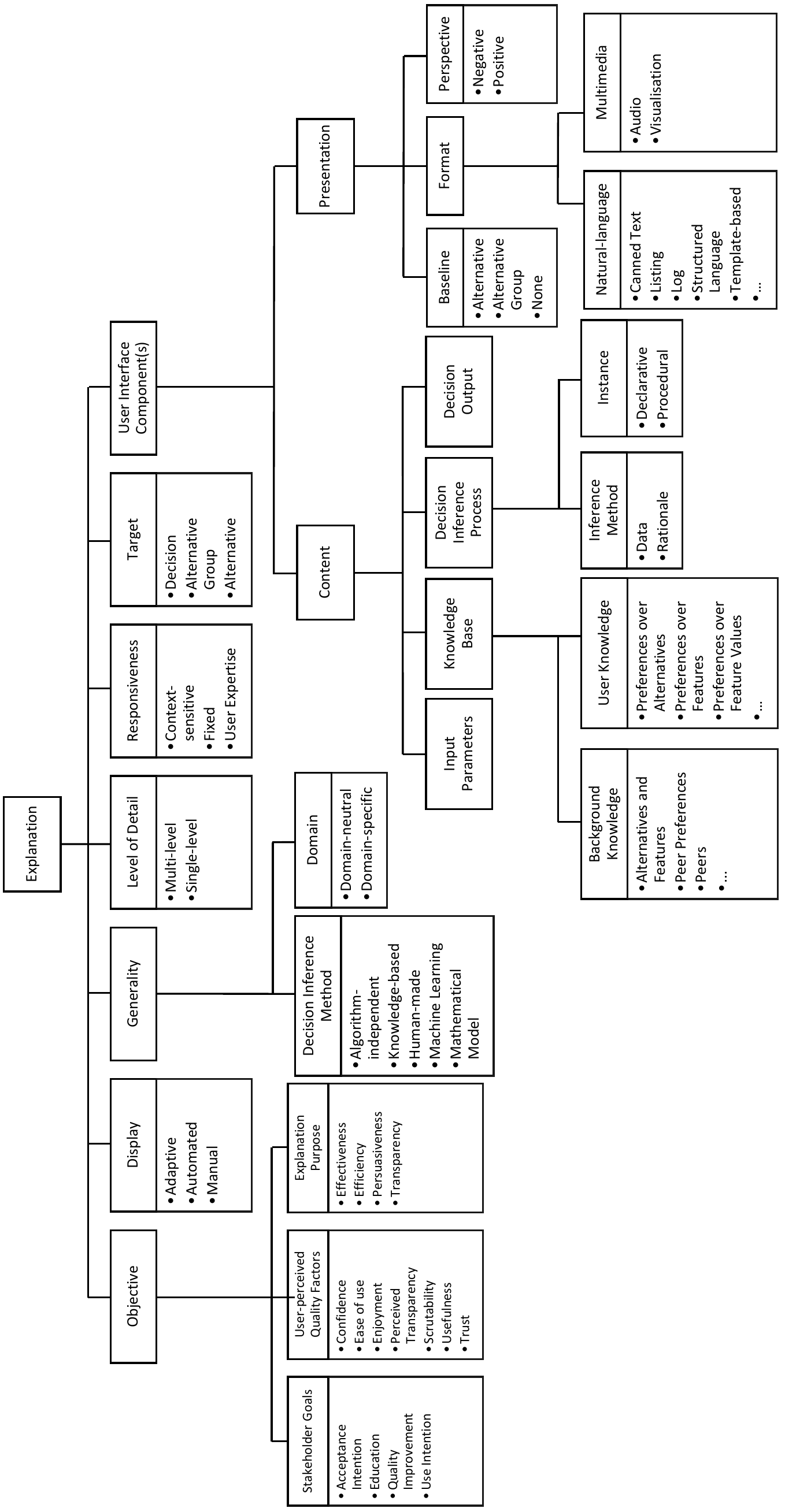}
	\caption{Explanation Taxonomy.}
	\label{fig:taxonomy}
\end{figure}

The comprehensive new taxonomy that we discuss next is based both on the primary studies that were investigated in our review as well as on the existing---and sometimes not fully compatible---taxonomies from the literature. The underlying idea of our taxonomy is that explanations that are presented to the user consist of one or more \emph{user interface components}. A component can be a justification in natural language, a histogram, or some other way of conveying information to the user. Our taxonomy, shown in Figure~\ref{fig:taxonomy}, includes both general facets that are associated with explanations and their generation approach as well as facets related to the content and the presentation of individual explanation components (referred to as user interface components), which collectively comprise an explanation.

\subsubsection{General Facets of Explanations} 

When designing an explanation approach, we should first determine \emph{what is the objective} of providing explanations to the users. Previous research work described a number of possible \emph{explanation purposes}, as discussed in Section~\ref{sec:rq2}. These possible purposes (e.g.\ transparency, efficiency, and trust) are often provided as a flat list, indicating that they are independent. In reality, as stated by Tintarev and Masthoff~\cite{296} and others, the possible purposes can, however, be related in different ways. Achieving transparency can, for example, constitute one of several factors that contribute to user trust in the system~\cite{491,744,630,Nilashi201670}. Moreover, trust is usually not even the ultimate goal from the perspective of the provider of the \ags{}, who might be interested in increasing the user intention to continue to use the system in the future. In the spirit of Jannach and Adomavicius's work~\cite{JannachAdomavicius2016purpose}, we distinguish between three levels of possible \emph{objectives} of explanations:

\begin{enumerate}[label=(\roman*),leftmargin=1.5\parindent]
	\item \emph{stakeholder goals}, e.g.\ to increase the user's intention to reuse the system;
	\item \emph{user-perceived quality factors}, which are system quality factors that can contribute to the achievement of the stakeholder goals (e.g.\ trust); and
	\item \emph{explanation purposes}, which are explanation-specific objectives that can contribute to achieving the objectives at the other levels.
\end{enumerate}

The key idea of having these three levels is that the choice and design of an explanation mechanism must be guided by the overarching goals that should be achieved with the explanations. Without clearly knowing the underlying motivations and in which ways the explanations are related to the quality perceptions of users, explanations might be added to a system in an arbitrary way. If, for example, the objective of the service provider is establish long-term customer relations (i.e.\ quality improvement), the service provided must optimise an adequate related quality factor at the user side (e.g.\ confidence). Based on the selected quality factor, the chosen explanation design should then support a corresponding purpose (e.g.\ effectiveness).

Based on previously proposed explanation purposes~\cite{296,Buchnan:chapter1984:ExplanationSurvey}, the different types of objectives can be further refined as shown in Figure~\ref{fig:taxonomy}. Additional details about the objectives at the lower levels are provided in Table~\ref{tbl:business-system-explanation}.
We highlight that, differently from Tintarev and Masthoff~\cite{296}, we do not consider satisfaction as a single objective, but split it into ease to use, enjoyment, and usefulness. As shown in the analysis of \textsc{evaluation} studies, these are investigated and evaluated separately, in addition to being possibly conflicting. Explanations may be provided to increase the perceived usefulness of the system, although this may compromise ease of use.
Moreover, note that we distinguish \emph{perceived transparency} (a user-perceived quality factor)  from \emph{transparency} (an explanation purpose). The information provided to a user may be perceived as detailing how a system works. However, such information may not necessarily match how the system \emph{actually} works.

\begin{table}
  \centering
	\footnotesize
  \caption{Description of the Objective Facet.}
	\label{tbl:business-system-explanation}
  \begin{tabularx}{\linewidth}{p{2cm} p{1.9cm} X}
    \toprule
    \textbf{Category} & \textbf{Property} & \textbf{Description} \\
		\midrule
    Stakeholder \par Goals & Acceptance Intention & Increasing the probability of users accepting the suggestion (referred to as purchase intention in e-commerce systems). \\ \cline{2-3}
		         & Education & Providing users with knowledge to make decisions in the system domain. \\ \cline{2-3}
		         & Use Intention & Increasing the probability of users using the system. \\ \cline{2-3}
		         & Quality\par Improvement & Improving the quality of user decisions (in terms of correctness) by detecting possible system flaws. \\
		\midrule
		User-perceived \par Quality Factors & Confidence & Being perceived as a system that helps users make good decisions, i.e.\ making users confident of the decision quality. \\ \cline{2-3}
		             & Ease of Use & Being perceived as easy to use. \\ \cline{2-3}
		             & Enjoyment & Being perceived as a system that brings enjoyment to users. \\ \cline{2-3}
								 & Perceived Transparency & Being perceived as transparent, i.e.\ a system that exposes its inner workings. \\ \cline{2-3}
				      	 & Scrutability & Being able to receive and use user feedback about the decision advice. \\ \cline{2-3}
				      	 & Usefulness & Being perceived as a useful system. \\ \cline{2-3}							
				      	 & Trust & Being perceived as a trustworthy system. \\
		\midrule
		Explanation\par Purposes & Effectiveness & Providing information to allow assessment of whether the suggested alternative is appropriate. \\ \cline{2-3}
		               & Efficiency & Providing information to help users make faster decisions. \\ \cline{2-3}
								   & Persuasiveness & Providing information to convince users that the suggested alternative is appropriate. \\ \cline{2-3}
							     & Transparency & Providing information to understand the inference logic of the advice-giving system. \\
	  \bottomrule
  \end{tabularx}
\end{table}

Our taxonomy includes five additional general facets. With the explanation \emph{Target} facet, we distinguish between situations in which the provided explanations refer to a single decision alternative, to a group of alternatives, or the decision output as a whole. The \emph{Generality} facet captures whether an explanation is generated using information that comes from a particular domain and, therefore, the explanation generation process is domain-specific and not general enough to generate explanations in other domains. Furthermore, generality also refers to the question whether the approach to generate explanations is tied to a specific underlying decision inference method or if it is able to generate explanations only considering specific decision inference methods. With the \emph{Responsiveness} facet, we distinguish between explanations that are adapted to the current user context and those that are not. In addition, the \emph{Level of Detail} indicates that explanations with more or less details can be provided, depending on a certain criterion, such as user expertise. More detailed explanations are more informative, but can require a higher cognitive effort from the user. To deal with this trade-off, there are approaches that provide multiple levels of detail, which can be explored by users as needed or displayed according to the current context. Finally, the \emph{Display} facet characterises what triggers an explanation to be displayed. The usual alternatives are manual (shown upon user request), automatic (always shown), or adaptive (depending on the context).

\subsubsection{User Interface Components}

The other facets of our taxonomy are concerned with aspects related to the \emph{content} and the \emph{presentation} of explanations. This part of the taxonomy is mainly orthogonal to the previously introduced facets. However, the design decisions that are made along the dimensions of these general facets can impact on these remaining choices. For example, if the explanation target is the decision as a whole, a baseline cannot be selected for a user interface component, because baselines make only sense when comparing alternatives.

Regarding the \emph{Content} facet, we identified four key types of information that can be presented in explanations, detailed as follows.

\begin{description}

	\item[Input Parameters.] User interface components that refer to the inputs that were provided for a particular decision problem, e.g.\ the set of symptoms that were entered in a medical decision support system or the current user mood in the case of a movie recommender system.
	
	\item[Knowledge Base.] User interface components that include information that resides in the knowledge base of an \ags{}. The provided information can be personalised, i.e.\ tailored to the specific user that receives the advise, or it can be background (or world) knowledge that is selected independent of the current user. Examples of the different types of knowledge are provided in Figure~\ref{fig:taxonomy}.
	
	\item[Decision Inference Process.] User interface components can also include information that is related to the system's internal process of determining the suggested alternatives. Such explanations can refer to a specific decision problem instance or provide general information about the internal inference method. In this latter case, the system can either explain the general idea behind the algorithm (e.g.\ recommendation of alternatives that similar users like) or the data it uses (e.g.\ use of users' shopping history to identify what they like). When the explanations are tied to the specific decision problem instance, the explanations can be procedural, i.e.\ describe the steps taken to reach a decision (e.g.\ rule trace), or declarative, providing information such as the confidence in the decision.

	\item[Decision Output.] Finally, user interface components can focus on the decision reasoning outcome and, for example, describe the particular features and feature values of the recommended and non-recommended alternatives. The explanation style \emph{Pros and Cons} presented in Table~\ref{tbl:orientationLabels} illustrates an explanation component that falls into this category.

\end{description}

Looking at the \emph{Presentation} facet of the taxonomy, there are three sub-facets. First, there are different ways of including none or multiple \emph{baselines} for comparison in the explanation. The baseline (or anchor) can be a single alternative to that recommended or a group of alternatives. Second, different output \emph{formats} can be chosen, as discussed previously in our review, such as using natural language or different types of visualisations. We list possible alternatives in their corresponding boxes. For example, canned text consists of a set of text segments that, when combined, form one sentence. Templates, in contrast, are usually almost complete sentences that must be completed with a set of arguments. Finally, the \emph{perspective} in which an explanation is presented can either be \emph{positive}, i.e.\ focusing on why an alternative is suitable for a user, or \emph{negative}, i.e.\ detailing why certain negative aspects of an alternative could be acceptable.

\subsection{Future Directions}

While a substantial amount of work has already been done in the context of explanations, a variety of open issues still need to be addressed. In this section, we give examples of such open research questions. These questions refer to several \emph{concerns} that need further investigation, ranging from questions related to the general objectives of explanations, over questions regarding the choice of the explanation content, to open methodological issues.

\paragraph{Understanding the Relationship among Stakeholder Goals, User-Perceived Quality Factors, and Explanation Purposes} In the previous section, we argued that the explanation purposes that are mentioned in the literature can be different in nature and some are only indirectly related to explanations. Correspondingly, we distinguished between Stakeholder Goals, User-Perceived Quality Factors, and Explanation Purposes in our taxonomy. These objectives are interrelated, e.g.\ transparency can have an impact on trust. Although one can form intuitive hypotheses about the relationships among these objectives, more systematic studies of these relationships still need to be done.

\paragraph{Selecting the Right Explanation Content} As shown in our taxonomy and in the discussion regarding explanation content, various types of information can be presented within explanations. Most of the user studies that we examined in this work compared largely different forms of explanations and there is no common \emph{baseline} explanation form that is used in many studies. Furthermore, the compared explanation forms often vary in many different aspects so that it is not possible to understand what content should be presented to which kind of users and when.
In the early years of expert systems, explanations often focused on the decision inference process and provided inference traces. However, it soon became clear that explanations that focus on \emph{why an alternative is adequate} are more helpful for users. Based on these considerations, we argue that explanation generation approaches should be further investigated in the future, being as independent as possible of the underlying decision inference process, thereby increasing the possibility to reuse approaches for different types of \agss{}. In our systematic review, only a small number (11) of primary studies consists of approaches that are detached from the underlying decision inference method. As a consequence, with the increasing adoption of complex machine learning methods, such forms of explanations will become increasingly more important in the future. The same holds for application domains in which the internal inference methods are confidential~\cite{536}. Nevertheless, algorithm-independent explanations might not be appropriate when systems autonomously make decisions regarding individuals. Regulations, such as the General Data Protection Regulation\footnote{\url{http://www.eugdpr.org/}}, have emerged to give citizens the right to go against algorithmic-based decisions that affect them. Consequently, the provision of \emph{transparent} explanations is required in these cases.

\paragraph{Investigating Fine-grained Details of Presentation Aspects} Various questions are also open with respect to the fine-grained details of how to present explanations to the users. Explanations can vary, for example, with respect to (i) their length; (ii) the adopted vocabulary if natural language is used; (iii) the presentation format, and so on. When explanation forms are compared in user studies that are entirely different in these respects, it is impossible to understand how these details impact the results. Therefore, more studies are required to investigate the impact of these variables. Such user studies could then provide a more solid foundation for the development of new explanation approaches. From all investigated studies, only two~\cite{509,256} of the proposed explanation forms were explicitly founded and motivated by a preliminary study. Many of the others studies might have used existing research results in the literature as foundation of their work. However, this could have been highlighted to justify design decisions associated with explanations.

\paragraph{Towards More Responsive Explanations} Intelligent \agss{} are adopted in a wide range of domains. In some of them (like health), decisions are more critical than in others (like movie recommendation). Therefore, it is important to better understand which forms of explanations are appropriate for which scenarios. When designing an explanation facility one should, for example, consider how much effort users might be willing to make in order to analyse explanations. Moreover, taking into account the user who is interacting with the system, and her background knowledge, is also relevant---as discussed in some of the analysed studies. However, only 16 explanation \textsc{techniques} and \textsc{tools}, i.e.\ 8\% of all studies of these types, aimed at providing explanations that are tailored to a given context or user expertise.

\paragraph{The Need for Adequate Objective Evaluation Protocols and Metrics} The most common way of assessing different explanation aspects, e.g.\ transparency or persuasiveness, is to use questionnaires that the participants fill out as part of or after an experiment, which is a well-established research instrument. Such studies rely on the participants' subjective perception of certain system or explanation aspects and on their behavioural intentions. Nonetheless, this research approach has limitations, as study participants may find it difficult to express to what extent they feel persuaded by a system or consider the system's explanations transparent, for example. A further limitation is that there are no standardised study designs or list of questionnaire items in the field.
Consequently, it is important that researchers develop a standardised set of evaluation protocols to measure certain aspects of explanations. Moreover, these protocols should rely on \emph{objective} measures, in addition to the subjective quality perception statements. One example of such a protocol is what we referred to as the \emph{explanation exposure delta}. However, this protocol can only be used to assess certain aspects, and more work towards a comprehensive evaluation framework for explanations is still required.

\subsection{Research Limitations}

The main goal of our systematic review is to develop a comprehensive understanding of what has been done in the field of explanations based on an unbiased selection and analysis of a large amount of primary studies. However, due to the systematic process of selecting the studies from a specified set of digital libraries, our survey does not cover all existing work on explanations. We generally selected widely used digital libraries as sources, which we assumed that would contain the largest number of relevant studies. An example of a comparably often cited study that was not retrieved in our search is that of Bilgic and Mooney~\cite{Bilgic:IUI-BP2005:ExplainingRecommendations}, because it was published in workshop proceedings. However, we believe that the number of relevant studies that were published only in a workshop and were not continued in a conference or journal paper is low. A few other relevant studies ~\cite{Nunes:ECAI2014:Explanations,Junker:AAAI2004:QuickXplain,Carenini:IJCAI2001:ExplanationStudy} were also not included in our review, because they are part of the the AAAI Digital Library\footnote{\url{http://www.aaai.org/Library/library.php}}, which unfortunately provides too limited search support to be usable for our survey. Nevertheless, the advantage of systematic reviews is that they can be further extended in future reviews, which can follow the specified procedure with other databases or in a future publication time range.

\section{Summary}\label{sec:sysRevConclusion}

The increasing trend of using software systems as advice givers and also making them more autonomous calls for approaches that allow systems to be able to \emph{explain} their decisions. Due to this trend, explanations increasingly receive more attention. However, a substantial amount of work has already been done in the field of explanations in \agss{}, mainly in the context of expert systems. Therefore, to have a comprehensive view of what has been done in this field, we presented the results of a systematic review of studies that proposed new techniques to generate explanations, described tools that include an explanation facility, or detailed the results of evaluation and foundational studies. A wide range of aspects associated with explanations were discussed in the 217 analysed studies.

We observed that most of the explanations provided in existing work consist of inference traces that were collected during the internal process of reasoning about which alternative(s) should be suggested to users. To be presented, such traces are often transformed into natural language statements. The adoption of this form of explanation was mainly due to the underlying inference method, which was some form of rule-based reasoning in most cases. The use of traces also explains the most frequent intended purpose of providing explanations, which is transparency (i.e.\ detail how a system reached a particular conclusion), so that users can trust the system. However, such traces are often not helpful for end users. Therefore, other forms of explanations were explored, such as those that focus on the features of the suggested alternatives and those that present visualisations rather than natural-language statements. This also led to the exploitation of explanations to persuade users or increase their satisfaction while interacting with the system. By analysing the results of studies which evaluated and compared explanations, we observed that there is a lack of standardised protocols and measurements to guide the study design, which makes a comparison of the obtained results difficult. However, a deeper analysis of results that were obtained mainly through user studies indicated divergence in terms of the reached conclusions. Consequently, there is a need for conducting further studies to identify the real pros and cons of explanations.

Based on our literature review, we proposed an explanation taxonomy, which covers a wide range of facets that can be used as a guideline by researchers when proposing and evaluating future approaches to generate explanations. Our taxonomy highlights the importance of specifying clear objectives as a key driver when designing new approaches, with objectives ranging from stakeholder goals to the specific explanation purposes. Moreover, we discussed open challenges that remain open, such as the understanding of the right explanation content according to different contexts and the subtleties of the explanations that may have an impact on the user, e.g.\ the used vocabulary. The identified challenges leave room for much research work that needs to be done in order for users to develop trust towards future \agss{} and autonomous systems.

Our review focused on explanations provided to end users, who need further information when receiving recommendations from \agss{}, with varying purposes. However, due to the increasing complexity of machine learning techniques, including those based on deep learning, providing explanations for data scientists to understand the outcomes of these techniques is becoming crucial, leading to what is generally referred to as explainable artificial intelligence. This broader topic is out of the scope of our work but should be further investigated in the future.


\section*{Acknowledgments} \label{sec:acknowledgments}
The authors would like to thank Michael Jugovac for carefully proofreading this paper. Ingrid Nunes also would like to thank for research grants CNPq ref. 303232/2015-3, CAPES ref. 7619-15-4, and Alexander von Humboldt, ref. BRA 1184533 HFSTCAPES-P.

\bibliographystyle{spmpsci}
\bibliography{explanations-systematic-review}

\textbf{Dr. Ingrid Nunes} is a Senior Lecturer at the Institute of Informatics, Universidade Federal do Rio Grande do Sul (UFRGS), Brazil, currently in a sabbatical year at TU Dortmund in Germany. She obtained her PhD in Informatics at the Pontifical Catholic University of Rio de Janeiro (PUC-Rio), Brazil. Her PhD was in cooperation with King's College London (UK) and University of Waterloo (Canada). She is the head of the Prosoft research group, and her main research areas are decision making and multi-agent systems.

\textbf{Dr. Dietmar Jannach} is a Professor of Computer Science at TU Dortmund, Germany and head of the department's e-services research group. Dr. Jannach has worked on different areas of artificial intelligence, including recommender systems, model-based diagnosis, and knowledge-based systems. He is the leading author of a textbook on recommender systems and has authored more than hundred technical papers, focusing on the application of artificial intelligence technology to practical problems.

\end{document}